\documentclass[10pt,conference]{IEEEtran}
\IEEEoverridecommandlockouts

\usepackage{xspace}
\usepackage{tcolorbox}
\usepackage{enumitem}
\usepackage[ruled]{algorithm2e}

\usepackage[numbers,sort]{natbib}
\usepackage[colorlinks=true,citecolor=blue,linkcolor=blue,urlcolor=blue,bookmarks=false]{hyperref}

\usepackage{booktabs} 
\usepackage{multirow}
\usepackage{graphicx}
\usepackage{tabularx}
\newcolumntype{C}[1]{>{\centering\arraybackslash}p{#1}}
\usepackage{colortbl}
\usepackage{graphicx}  
\usepackage{float}  
\usepackage{subfig}
\usepackage{makecell} 
\usepackage{color}
\usepackage{arydshln} 
\usepackage{caption}
\usepackage{wrapfig}
\usepackage{colortbl}
\usepackage{amsthm} 
\usepackage{amsmath}
\usepackage{adjustbox}
\usepackage{threeparttable}
\usepackage{stfloats}
\usepackage{pifont}
\usepackage{xurl}
\usepackage{amsfonts}

\theoremstyle{definition}

\usepackage{microtype}
\newcommand{\ie}[0]{\textit{i.e.,}\xspace}
\newcommand{\eg}[0]{\textit{e.g.,}\xspace}

\newcommand{\tool}{\textsc{DocsChisel}\xspace}

\newcommand{\todo}[1]{\textcolor{black}{#1}}
\newcommand{\ly}[1]{\textcolor{black}{#1}}

\def\BibTeX{{\rm B\kern-.05em{\sc i\kern-.025em b}\kern-.08em
    T\kern-.1667em\lower.7ex\hbox{E}\kern-.125emX}}
\begin{document}

\title{\tool: Adaptive Tool Documentation Optimization Framework for LLM Agents}

\author{\IEEEauthorblockN{
You Lu\IEEEauthorrefmark{1},
Kun Zhang\IEEEauthorrefmark{1}, 
Bihuan Chen\IEEEauthorrefmark{1},
Xin Peng\IEEEauthorrefmark{1}}
\IEEEauthorblockA{\IEEEauthorrefmark{1}College of Computer Science and Artificial Intelligence, Fudan University, China}}

\maketitle

\begin{abstract}
Large language models (LLMs) increasingly rely on external tools to accomplish complex real-world tasks, making tool documentation a critical grounding resource for LLM agents. Existing studies mainly focus on improving the \ly{tool-use} capabilities of LLM agents, while largely treating tool documentation as a fixed input. Although several recent works attempt to optimize tool documentation through rewriting or compression, little is known about how the information contained in tool documentation affects agent performance across different settings. 

To bridge this gap, we conduct a large-scale empirical study on tool documentation for LLM agents. Our study~reveals~substantial heterogeneity in the information \ly{fields} provided by existing~tool documentation. Moreover, the effectiveness of different information fields is highly dependent on the task domain, LLM backbone, and agent paradigm, indicating that no fixed tool documentation can consistently generalize across diverse agent~settings.

Motivated by these findings, we propose \tool, an~adaptive tool documentation optimization framework for LLM agents. \tool analyzes failed execution traces of a target LLM agent to identify documentation-related issues, and iteratively optimizes tool documentation by adding, removing, and refining information fields for each tool.  We evaluate \tool against two state-of-the-art baselines, \ie \textsc{EasyTool} and \textsc{DRAFT}. Experimental results show that \tool improves the task success \ly{rate} of LLM agents by \ly{95.89\%} over the original tool documentation and by \ly{75.15\%}, on average, over existing baselines, while incurring limited optimization time and token overhead.

\end{abstract}

\section{Introduction}\label{sec:intro}
Large language models (LLMs)~\cite{chatgpt, gemini} have demonstrated strong capabilities across diverse tasks~\cite{hou2024large, glm2024chatglm, huang2023chatgpt}, such as code generation, question answering, and data analysis. To execute real-world tasks, LLMs increasingly rely on external tools, \eg web browsers, code interpreters, and file systems, leading to the emergence of \ly{tool-use} LLM agents (referred to as LLM agents hereafter)~\cite{yao2022react, autogpt2023, yang2024swe, schick2023toolformer, apibench, toolllm}. Emerging LLM agent ecosystems, including agent development platforms~\cite{coze, dify}, tool hosting platforms~\cite{rapidapi, openapihub}, and agent evaluation suites~\cite{toolbench, apibank}, provide large collections of reusable tools and standardized tool interfaces. These tools are typically accompanied by tool documentation that includes several information fields about their functionalities, invocation constraints, and other details, helping LLM agents understand and invoke tools correctly.

Traditional API documentation has long been studied in software engineering, primarily to improve its readability, completeness, and usability for developers~\cite{dekel2009improving, de2009automatic, maalej2013patterns, nybom2018systematic, shi2011empirical, khan2021automatic, piccioni2013empirical, nahar2022collaboration, subramanian2014live}. However, tool documentation for LLM agents differs in both audience and content. The documentation consumers are LLM agents that must infer tool functionality and generate executable tool \ly{invocations} directly from in-context documentation. Meanwhile, the documentation emphasis shifts from general API understanding to task-oriented tool use, requiring information such as usage conditions, parameter semantics, invocation constraints, and output interpretation. As such documentation guides runtime tool selection and invocation, ambiguity or missing information can lead to incorrect tool \ly{invocations} and cascading task failures. Thus, tool documentation should be viewed not only as a human-readable reference, but also as a critical grounding resource whose quality heavily affects the task success rate of~LLM~agents~\cite{hsieh2023tooldoc, easytool, toolllm}.

\textbf{Literature.}
Despite the importance of tool documentation, existing studies~\cite{mavroudis2024langchain, toollens, zhuang2026losemb, toole, taubench, anytoolbench, apibank, shortcutsbench, shi2025tool}~mainly focus on adding or improving \ly{tool-use} capabilities of~LLM agents from the perspectives of tool construction, tool~retrieval, tool invocation, and evaluation. While these studies~have~significantly advanced the development of LLM~agents, they~usually treat tool documentation as a fixed given input,~without systematically examining how documentation itself~affects~LLM agent behavior and task success \ly{rate}. Only a few recent studies \cite{jtpro, play2prompt, liu2025toolscope, easytool, guo2026learning, qu2025exploration} explicitly optimize tool documentation for LLM agents. For instance, \textsc{EasyTool}~\cite{easytool} standardizes diverse tool documentation into concise tool instructions with a unified template, thereby reducing the context overhead~caused by documentation, whereas \textsc{DRAFT}~\cite{qu2025exploration} \ly{iteratively}~refines tool documentation based on \ly{outputs of external tools}. However, these approaches mostly optimize tool documentation within existing information fields through rewriting, correction, standardization, or compression. They pay less attention to whether the documentation fields themselves are sufficient, redundant,~or~suitable for different LLM agents.

In practice, tool documentation across LLM agent ecosystems may differ substantially in the information fields~it~provides. \ly{However, it remains unclear whether the utility~of~these information fields varies across agent settings. For example, do agents powered by less capable LLM backbones benefit more from explicit usage examples? Do multi-agent paradigms place greater demands on usage guidance and invocation constraints?} These observations raise a fundamental question, i.e., \textit{what information fields are provided by existing tool documentation, and how do these information fields affect the task success rate of LLM agents under different agent settings?}

\begin{table*}[!t]
  \centering
  \caption{Overview of the Collected Tool-Use Datasets}
  \label{tab:benchmark_overview}
  \begin{adjustbox}{width=0.9\textwidth}
  \begin{tabular}{lcl}
    \toprule
    \textbf{Dataset} & \textbf{\#Tools} & \textbf{Description} \\
    \midrule
    WorkBench~\cite{workbench} & 26  & A tool-use LLM agent benchmark with sandboxes, realistic office tasks, databases, and executable tools. \\
    API-Bank~\cite{apibank} & 2,211  & A comprehensive evaluation benchmark with diverse APIs and multi-turn dialogues for LLM agents. \\
    ToolLLM~\cite{toolllm} & 16,464  & A large-scale suite of real-world APIs for evaluating multi-step tool invocations by LLM agents. \\
    APIBench~\cite{apibench} & 1,645  & A massive API corpus for studying LLM connectivity to thousands of remote APIs. \\
    ToolAlpaca~\cite{ToolAlpaca} & 426  & A simulated dataset with diverse tool APIs and structured tool documentation for training LLMs. \\
    AnyToolBench~\cite{anytoolbench} & 13  & A benchmark for hierarchical, large-scale tool calling with self-reflective LLM agents. \\
    ToolBench~\cite{toolbench} & 232 & A benchmark for evaluating open-source LLMs on tool manipulation and invocation tasks. \\
    ShortcutsBench~\cite{shortcutsbench} & 1,414  & A large-scale benchmark based on Apple Shortcuts, containing real APIs, user queries, and action sequences. \\
    WildToolBench~\cite{yu2026wildtoolbench} & 1,600  & A tool-use benchmark grounded in real-world user behavior patterns and realistic multi-step scenarios. \\
    SWE-bench~\cite{jimenez2024swe,yang2024swe} & 15  & A software engineering benchmark requiring LLM agents to resolve real GitHub issues via repository tools. \\
    CRMArena~\cite{crmarena} & 27  & A domain-specific benchmark for professional CRM tasks in realistic enterprise environments. \\
    $\tau$-bench~\cite{taubench} & 28  & A benchmark for interactions among tools, agents and users with domain-specific tools and dialogues. \\
    TooLe~\cite{toole} & 390  & A meta-tool benchmark for deciding whether to invoke tools and selecting appropriate tools. \\
    ToolLens~\cite{toollens} & 464  & A tool-retrieval dataset designed for multi-tool scenarios, emphasizing complete and diverse tool selection. \\
    \bottomrule
  \end{tabular}
  \end{adjustbox}
\end{table*}

\textbf{Empirical Study.} To understand tool documentation for LLM agents and its impact on task success rate, we conduct a large-scale empirical study across 14 tool-use datasets collected from LLM agent ecosystems. Specifically, we investigate their information field composition, and evaluate the task success rate of two different LLM agent paradigms (\ie ReAct~\cite{yao2022react} and Multi-Agent~\cite{wu2024autogen}) on WorkBench~\cite{workbench}, using three different LLM backbones (\ie GPT-4o~\cite{gpt4o}, GLM-5~\cite{glm5} and Claude Haiku 4.5~\cite{claudehaiku}). We design the following~research questions.

\begin{itemize}[leftmargin=*]
    \item \textbf{RQ1 Composition Analysis.} What information fields are commonly contained in different tool \ly{documentation}?
    \item \textbf{RQ2 Impact Analysis.} How do different information fields affect the task success rate of LLM agents across different task domains, LLM backbones, and agent paradigms?
\end{itemize}

Our results reveal that existing tool documentation~in~agent ecosystems exhibits substantial heterogeneity in field composition, while different information fields have different impacts on task success rate of LLM agents across task domains, LLM backbones, and agent paradigms. 
\ly{Specifically,~we~identify~17 information fields, among which only \textit{tool name} and \textit{functionality description} are provided by all 14 datasets, while \textit{usage guidance} and \textit{invocation constraint} appear in only two datasets. Moreover, the same information field may have different or even opposite effects under different agent settings. On average, adding or removing a single information field changes the task success rate of LLM agents by \todo{6.34} percentage points in absolute terms.} These findings demonstrate that fixed tool documentation cannot generalize well across different LLM agent settings and should be optimized adaptively.

\textbf{Our Approach.}
Motivated by these insights, we propose \tool, an adaptive tool documentation optimization framework for LLM agents. Specifically, given a target LLM agent,~\tool first executes representative tool-use queries with the original tool documentation, and collects~failed agent execution traces. Then, \tool analyzes these failed~traces to identify documentation-related issues, and uses LLMs to iteratively revise tool documentation. Instead~of~simply rewriting tool documentation into a shorter or more general form, \tool optimizes both the content and structure of tool documentation by adding, removing, or refining~information fields for each tool, thereby generating adaptive tool documentation for different LLM agent settings to improve both tool invocation correctness and task success rate.

\textbf{Evaluation.}
We implement a prototype of \tool, and conduct experiments \ly{on \ly{74} tools across 9 task domains selected from WorkBench~\cite{workbench} and API-Bank~\cite{apibank}} to demonstrate its effectiveness and efficiency, compared with two state-of-the-art baselines, \ie \textsc{EasyTool}~\cite{easytool} and DRAFT~\cite{qu2025exploration}. Experimental results~demonstrate that \tool consistently improves task success rate under diverse agent settings~by~\ly{95.89\%}~compared with using the original tool documentation, and achieves an average of \ly{75.15\%} improvement compared with baselines. \ly{Besides, \tool maintains practical optimization efficiency, requiring \ly{12.65} minutes per tool, on average, to optimize the tool documentation for LLM agents.}

\textbf{Contribution.} This work makes the following contributions.

\begin{itemize}[leftmargin=*]
    \item We conduct a systematic empirical study on tool documentation for LLM agents, revealing the varying effectiveness of different information fields across different agent settings.
    \item We design and implement an adaptive tool documentation optimization framework, \tool, for LLM agents, thereby improving the tool invocation correctness and task success rate of LLM~agents.
    \item We conduct extensive experiments to demonstrate the  effectiveness and efficiency of \tool.
\end{itemize}


\section{Empirical Study}\label{sec:empirical}

To systematically investigate the two research questions~introduced in Sec.~\ref{sec:intro}, we present a large-scale empirical study on the information field composition of different tool documentation, and evaluate the task success rate of different LLM agent settings with different tool documentation.

\subsection{Study Design}
\textbf{Documentation Collection.}
We collect tool documentation from 14 representative datasets shown in~Table~\ref{tab:benchmark_overview}, covering diverse tool-use scenarios, \eg workflow execution, mobile automation, and software engineering tasks. We select these datasets based on the following criteria: \textit{(1) providing rich tool collections and corresponding tool documentation, (2) covering different task domains and execution environments, (3) having strong community influence and being widely used in recent studies on LLM agents.} In~total, we collect \ly{24,955} tools along with their corresponding tool~documentation.

\textbf{LLM Selection and Agent Paradigm Setting.}
To evaluate the effectiveness of tool documentation under different LLM agent settings, we select three representative LLMs as agent~backbones, \ie GPT-4o~\cite{gpt4o}, GLM-5~\cite{glm5}, and Claude Haiku~4.5~\cite{claudehaiku}, covering diverse model families widely used in recent LLM agents. We further instantiate two agent paradigms, \ie a ReAct agent implemented with LangChain~\cite{mavroudis2024langchain}, and a Multi-Agent system implemented with AutoGen~\cite{wu2024autogen}. 

\textbf{Evaluation Metric.}
We use \textit{task success rate} ($\text{TS}$)~to~measure whether an LLM agent successfully completes user~queries. Given a test suite containing tool-use queries $\mathcal{Q}$, for each~query ${q_i} \in \mathcal{Q}$, we denote its execution outcome as $s(q_i)$. $s(q_i) = 1$ indicates that the task is correctly completed, and $s(q_i) = 0$ otherwise. We compute the task success rate as $\text{TS} = \frac{1}{|\mathcal{Q}|}\sum_{i=1}^{|\mathcal{Q}|}\mathbb{I}(s(q_i) = 1)$, where $\mathbb{I}(\cdot)$ is the indicator function.

\textbf{RQ Setup.} For \textbf{RQ1}, we manually inspect the tool documentation from the 14 datasets we collected in Table~\ref{tab:benchmark_overview}. Because these datasets use different formats and inconsistent~information field names, three authors independently extracted candidate information fields, and then aligned semantically equivalent information through discussion using unified names following the most commonly used field names in the collected datasets. Disagreements were resolved by consensus to reduce individual bias. Based on the final aligned taxonomy, we report the~coverage ratio of each type of information field across~datasets.

\ly{For \textbf{RQ2}, we conduct the empirical evaluation using the tool documentation provided in WorkBench~\cite{workbench}, which contains 26 tools from five task domains, \ie data analysis, email management, calendar management, project management, and customer relationship management (CRM).} We modify the tool documentation by removing or adding one information field at a time while keeping all remaining fields unchanged according to the taxonomy derived from \textbf{RQ1}. Specifically, if the target information field already~exists in the tool documentation,~we remove it to measure the impact of missing information.~Otherwise, we manually supplement the information field based on the original documentation and dataset-provided metadata, \eg tool code and correct execution outcome, to measure the impact of additional information. \ly{Based on these documentation variants, we conduct controlled experiments to investigate whether the effectiveness of information fields remains consistent across different task domains, LLM backbones, and agent~paradigms.}  

First, to study whether the impact varies across task domains, we use GPT-4o with~the ReAct agent, and~evaluate~documentation variants on the~five task domains in WorkBench. Second, to study whether the impact varies across LLM~backbones, we focus on the data analysis domain and the ReAct agent, and repeat the experiments with three different LLM backbones, \ie GPT-4o, GLM-5, and Claude Haiku 4.5. Third, to study whether the impact varies across agent paradigms, we fix the data~analysis domain and GPT-4o, and compare the results under~the~ReAct and Multi-Agent paradigms. In all experiments, we keep~the query set, prompt template, and execution environment unchanged within each~comparison.


\begin{table}[t]
    \centering
    \caption{Prevalence of Information Fields across Datasets}
    \label{tab:field_prevalence}
    \begin{adjustbox}{width=\linewidth}
    \begin{tabular}{llc}
    \toprule
    \textbf{Information Field} & \textbf{Brief Description} & \textbf{Prevalence} \\
    \midrule
    Tool Name (TN) & Invocation identifier of target tool & 14/14 \\
    Functionality Description (FD) & Tool purpose and capability & 14/14 \\
    Task Domain (TD) &Application scenario or category  & 6/14 \\
    Input Parameter Name (IPN) & Input argument identifier & 13/14 \\
    Input Parameter Description (IPD) & Input argument semantics & 12/14 \\
    Input Parameter Type (IPT) & Input data types or formats & 13/14 \\
    Default Parameter Value (DPV) & Default input parameter values & 3/14 \\
    
    Parameter Required Flag (PRF) & Required argument indicator & 10/14 \\
    Parameter Optional Flag (POF) & Optional argument indicator & 7/14 \\

    Return Parameter Name (RPN) & Output argument identifier &4/14 \\
    Return Parameter Description (RPD) & Output argument semantics &5/14 \\
    Return Parameter Type (RPT) & Output data types or formats & 5/14\\

    Code Implementation (CI) & Tool implementation snippets & 5/14 \\
    Usage Example (UE) & Concrete invocation examples & 4/14 \\
    Response Template (RT) & Expected tool response format & 4/14\\
    Usage Guidance (UG) & Tool usage recommendations & 2/14 \\
    Invocation Constraint (IC) & Invocation conditions or restrictions & 2/14 \\
    \bottomrule
    \end{tabular}
    \end{adjustbox}
\end{table}

\subsection{Composition Analysis (RQ1)}\label{sec:rq1}
After manually analyzing the collected tool documentation, we observe that tools within the same dataset usually follow a consistent documentation convention, \ie tools in the same dataset tend to contain the same set of information fields. However, such conventions vary across datasets, leading to different field compositions in existing tool documentation.

We identify 17 common information fields, and report their prevalence across the 14 datasets. As shown in Table~\ref{tab:field_prevalence}, \ly{fields describing tool identity and input schemas are widely adopted. \textit{Tool name} and \textit{functionality description} appear in all datasets, while \textit{input parameter name}, \textit{input parameter description}, and \textit{input parameter type} appear in most datasets. In contrast, fields supporting task-oriented tool use and result interpretation, such as \textit{default parameter value}, \textit{return parameter name}, \textit{return parameter type}, \textit{usage example} and \textit{response template}, are provided much less consistently. \textit{Usage guidance} and \textit{invocation constraint} appear in only two datasets.}

    \textit{\textbf{Findings.}} Existing tool documentation shows cross-dataset heterogeneity in information field composition. Although tools within the same dataset follow a consistent convention, different datasets provide different sets of information fields.


\subsection{Impact Analysis (RQ2)}\label{sec:rq2}

\begin{figure*}
    \centering
    \subfloat[Data Analysis+GPT+ReAct]{
        \includegraphics[width=0.32\textwidth]{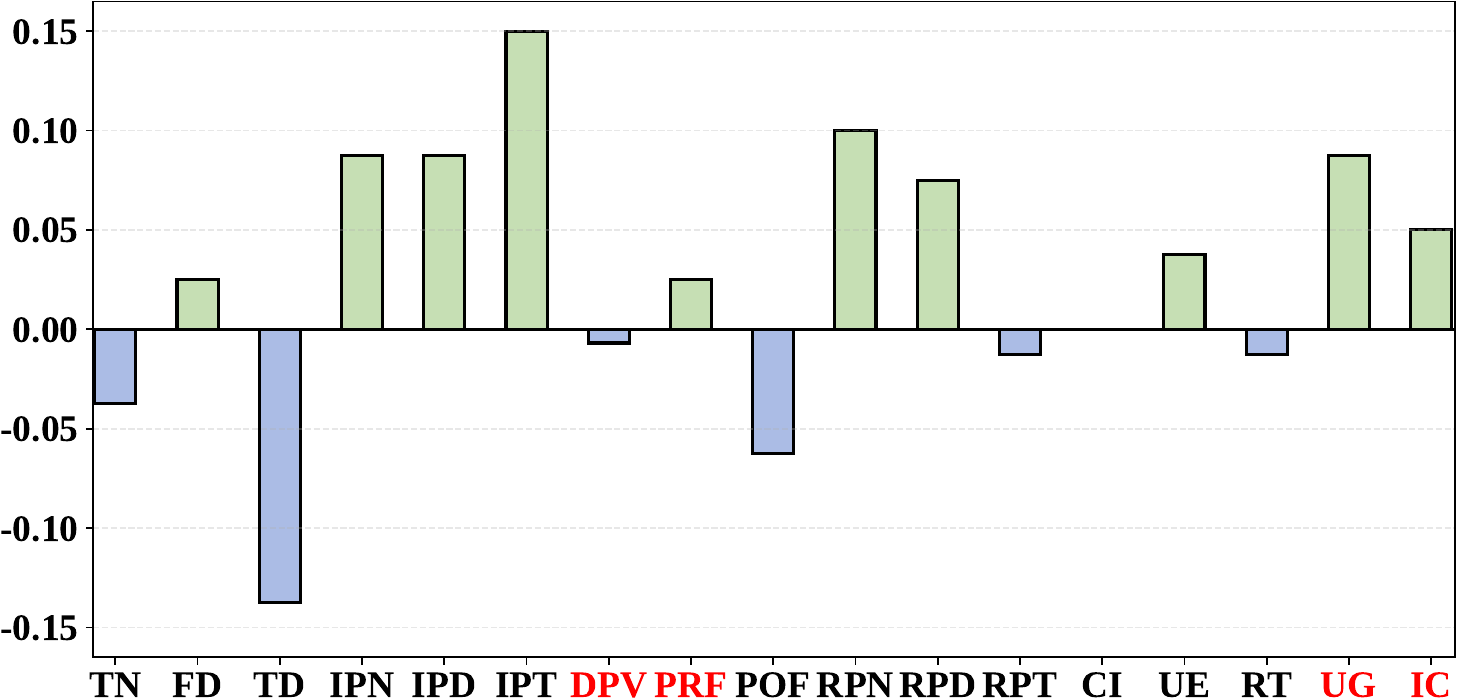}
        \label{img:DataAnalysis+GPT+ReAct}
    }
    \subfloat[Email+GPT+ReAct]{
        \includegraphics[width=0.32\textwidth]{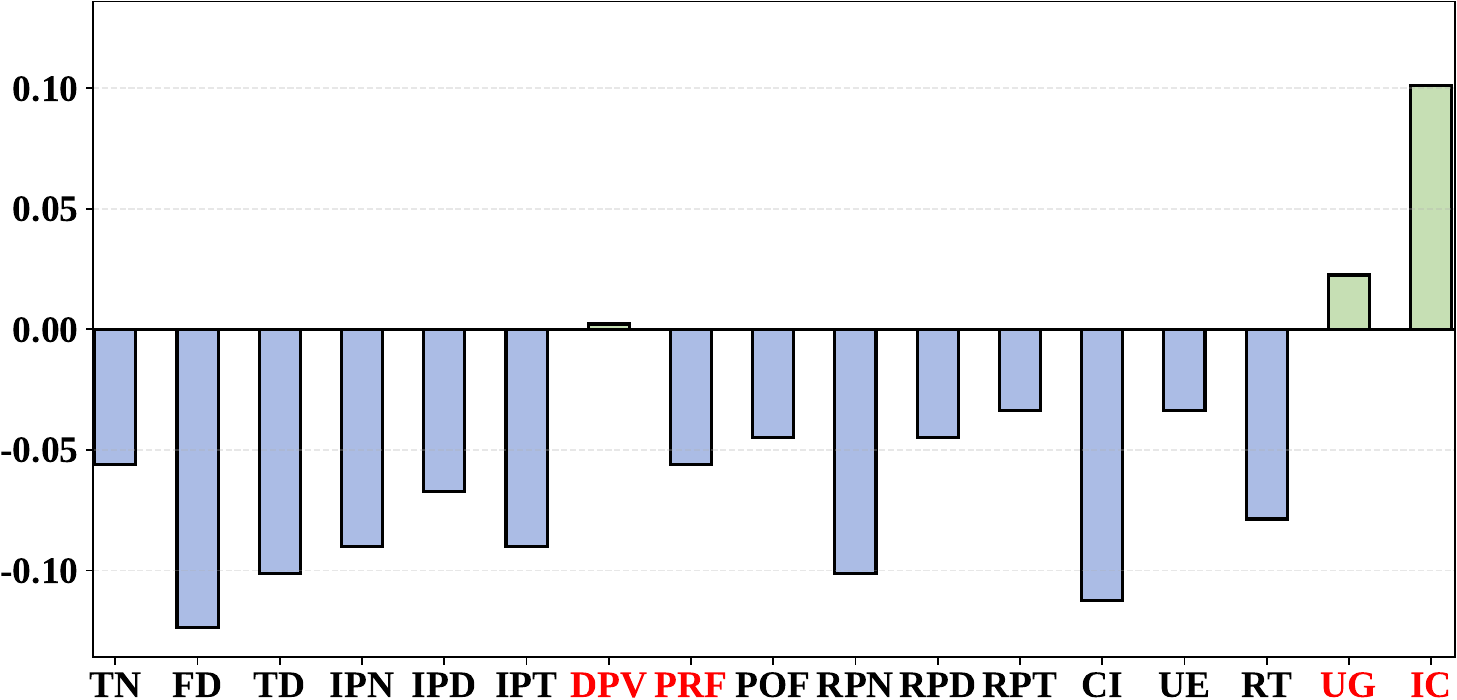}
        \label{img:EmailManagement+GPT+ReAct}
    }
    \subfloat[Calendar+GPT+ReAct]{
        \includegraphics[width=0.32\textwidth]{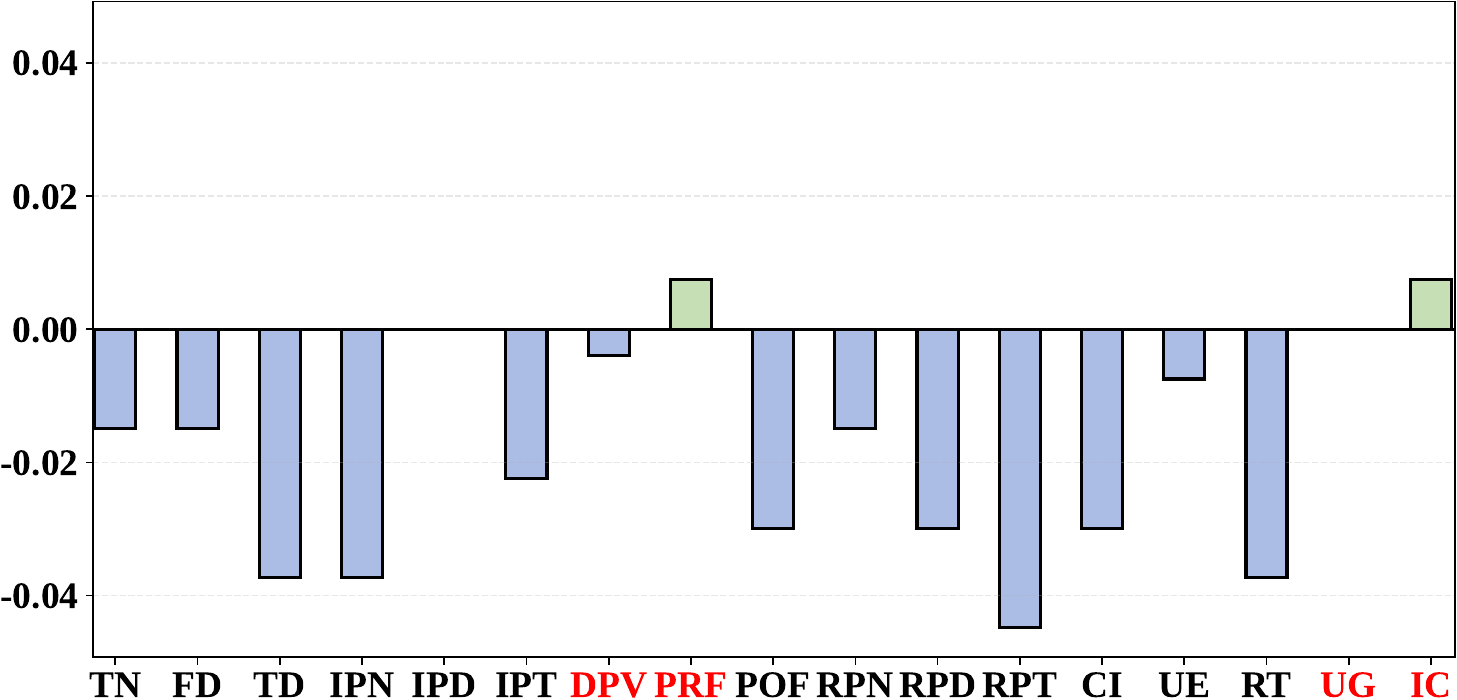}
        \label{img:CalendarManagement+GPT+ReAct}
    }

  \subfloat[Data Analysis+GLM+ReAct]{
      \includegraphics[width=0.32\textwidth]{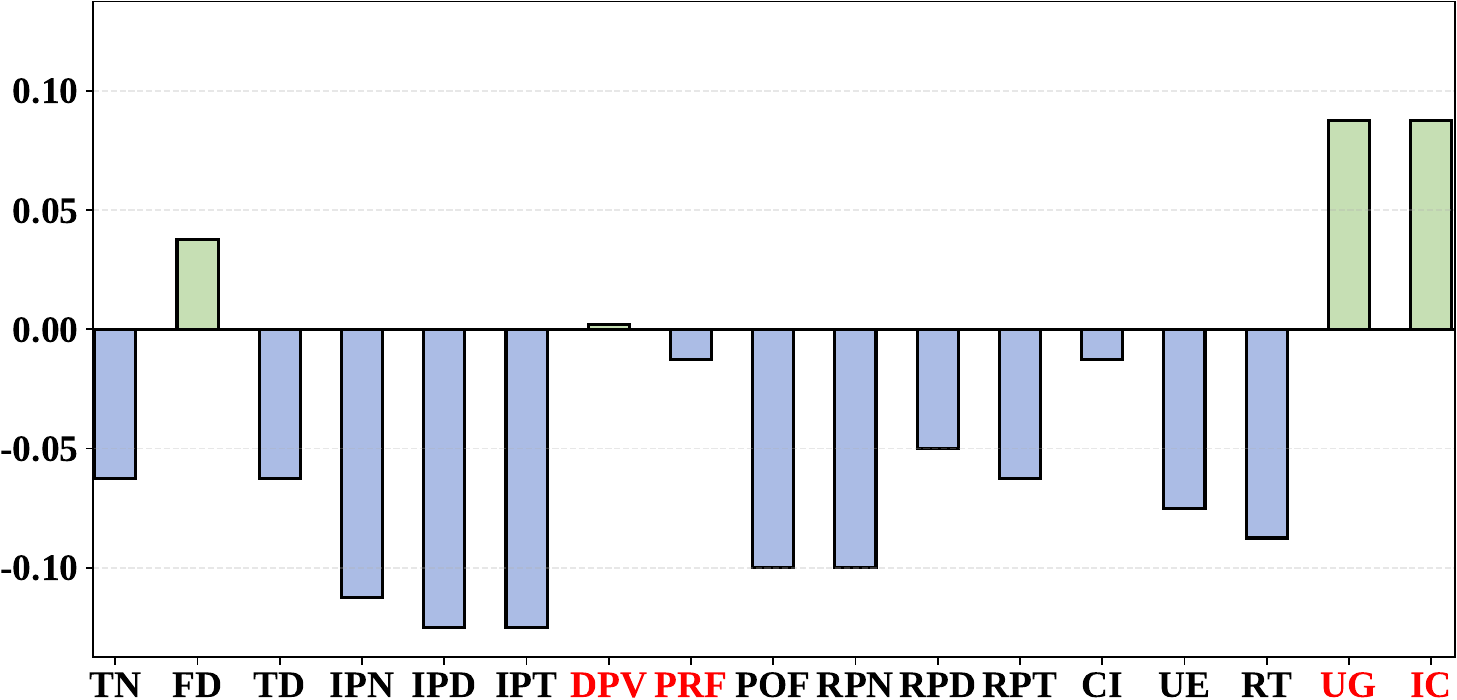}
      \label{img:DataAnalysis+GLM-5+ReAct}
  }
  \subfloat[Data Analysis+Claude+ReAct]{
    
    \includegraphics[width=0.32\textwidth]{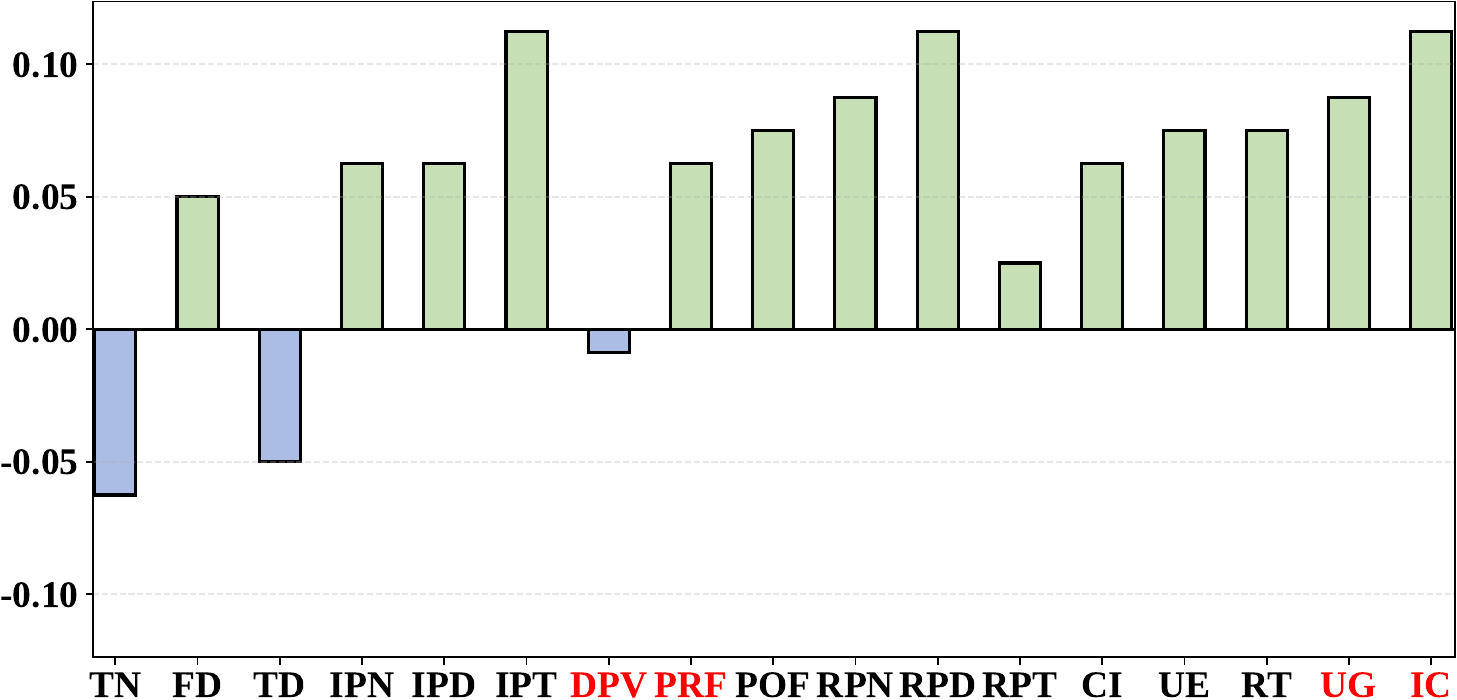}
    \label{img:DataAnalysis+Claude+ReAct}
}
\subfloat[Data Analysis+GPT+Multi-Agent]{
  
    \includegraphics[width=0.32\textwidth]{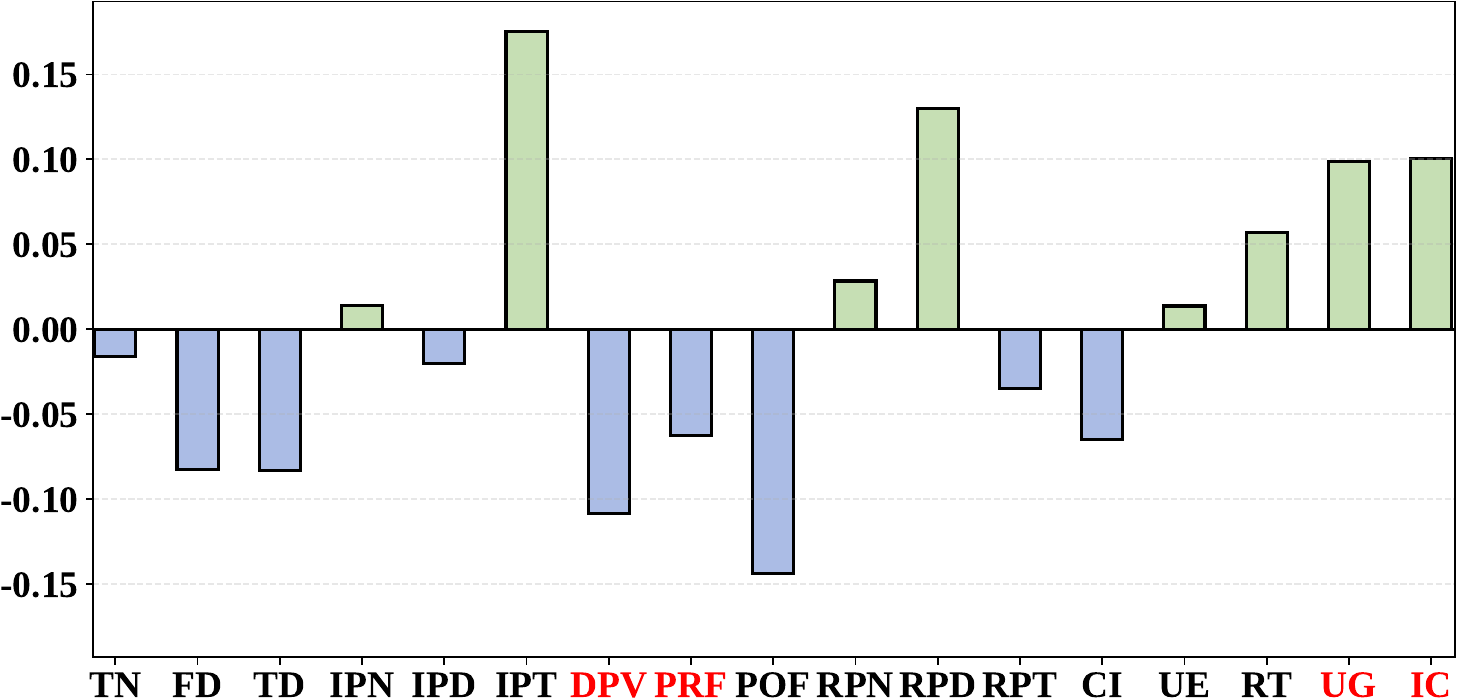}
    \label{img:DataAnalysis+GPT+Multi-Agent}
}
\caption{Impact of Information Fields on Task Success Rate across Task Domains, LLM Backbones, and Agent Paradigms}
\label{fig:field_impact}
\end{figure*}

Fig.~\ref{fig:field_impact} reports the change in $\text{TS}$ after adding or removing one information field from the original tool documentation. Black x-axis labels denote removed fields, while red labels denote added fields.
As shown in Fig.~\ref{img:DataAnalysis+GPT+ReAct}-\ref{img:CalendarManagement+GPT+ReAct}, \ly{under GPT-4o with ReAct, the evaluated fields exhibit both consistent~and domain-dependent effects. Among the 17 fields, only \todo{6} fields show the same~effect direction across all three domains.~Removing \textit{tool name} (TN), \textit{task domain} (TD), \textit{return parameter type} (RPT), \textit{response template} (RT), or \textit{parameter optional flag}~(POF)~consistently decreases $\text{TS}$, whereas adding \textit{invocation constraint}~(IC)~consistently improves it. For example, removing TD decreases~$\text{TS}$ by \todo{3.75} to \todo{13.75} percentage points, while adding IC improves it by \todo{0.75} to \todo{11.25} percentage points. In contrast, the remaining fields show domain-dependent effects. Adding \textit{parameter~required flag} (PRF), for instance, changes $\text{TS}$ by \todo{+2.50}, \todo{-5.61}, and \todo{+0.75} percentage points in data analysis, email management, and calendar management, respectively. Due to space limitations, results for project management and customer relationship management are provided on our website~\cite{website}.}

\ly{The effects also vary across LLM backbones. As shown in Fig.~\ref{img:DataAnalysis+GPT+ReAct}, Fig.~\ref{img:DataAnalysis+GLM-5+ReAct}, and Fig.~\ref{img:DataAnalysis+Claude+ReAct}, when fixing the domain and agent paradigm, \todo{12} of the 17 fields exhibit different effect directions across GPT-4o, GLM-5, and Claude Haiku 4.5. For example, removing POF decreases $\text{TS}$ for GPT-4o and GLM-5, but improves it by 7.5 percentage points for Claude Haiku 4.5. This indicates that different LLM backbones rely on different documentation information. A similar variation is observed across agent paradigms. As shown in Fig.~\ref{img:DataAnalysis+GPT+ReAct} and Fig.~\ref{img:DataAnalysis+GPT+Multi-Agent}, replacing ReAct with Multi-Agent reverses the effects of several fields, including \textit{functionality description} (FD), \textit{input parameter description} (IPD), \textit{parameter required flag} (PRF), and RT. Therefore, although some fields provide stable benefits or harms, the effectiveness of fields depends on the task domain, LLM backbone, and agent paradigm.}

    \textit{\textbf{Findings.}} Different information fields in tool documentation have different impacts on task success rate of LLM agents across task domains, LLM backbones, and agent paradigms.

\subsection{Empirical Insights}
The empirical study provides two insights for tool documentation optimization. First, no fixed documentation convention can consistently generalize across different LLM agent settings, since the effectiveness of information fields may vary across task domains, LLM backbones, and agent paradigms. Second, an effective tool documentation optimizer should adaptively adjust the information field composition based on the LLM agent setting according to different execution feedback.

\section{Methodology}\label{sec:approach}

Motivated by the empirical insights in Sec.~\ref{sec:empirical}, we propose \tool, an adaptive framework for optimizing tool documentation for LLM agents. Given a target LLM agent, a set of tools with their original documentation, and a set of queries with execution ground truth, \tool generates optimized tool documentation for each tool under the target agent setting. The key idea is to perform tool documentation optimization at the information-field level \ly{for each tool under the guidance of failed execution traces of the target agent.} 

\subsection{Approach Overview}\label{sec:approach_overview}
\begin{figure}[t]
    \centering
    \includegraphics[width=\linewidth]{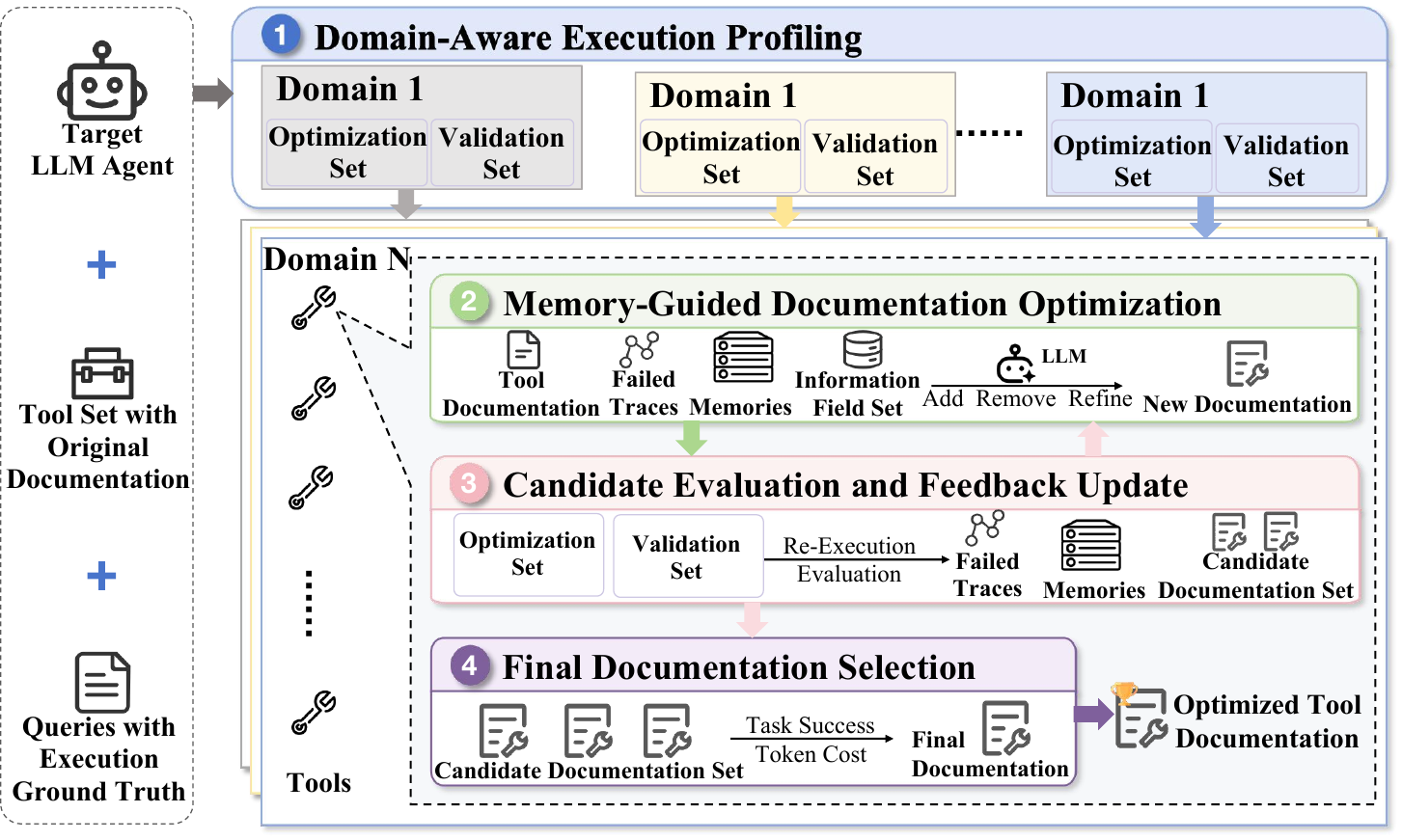}
    \caption{Approach Overview of \tool}
    \label{fig:overview}
\end{figure}

Fig.~\ref{fig:overview} shows the approach overview. First, \tool performs domain-aware execution profiling with the original tool documentation (Sec.~\ref{sec:profiling}). In this stage, \ly{when the target agent is equipped with tools from multiple task domains, \tool organizes the tools and queries by task domain. Within each domain, it associates each tool with the queries whose execution requires that tool, and partitions these tool-associated queries and their execution ground truth into optimization and validation sets. It then executes the target agent to collect traces on each tool's optimization set and obtain its initial task success rate on the corresponding validation set.}

Then, \tool enters an iterative optimization loop for each tool in the current task domain.  At each iteration, \tool performs memory-guided documentation optimization for the target tool using the queries associated with that tool in the optimization set (Sec.~\ref{sec:generation}). Given the original tool documentation, the latest failed execution traces collected in the optimization set, \ly{the field optimization memories of the current domain, and the information field set}, \tool uses an LLM-based exploration process to generate new tool documentation, including adding missing information fields, removing distracting or redundant fields, and refining ambiguous or incomplete fields. Each generated tool documentation is then evaluated through agent re-execution to update the feedback (Sec.~\ref{sec:feedback}). \ly{\tool first executes the target agent with the generated tool documentation on the optimization set to obtain new failed traces, and updates the field optimization memories for the next round of tool documentation generation. Then, it evaluates the generated tool documentation on its validation set to update the candidate documentation set. Once the maximum number of optimization iterations for the target tool is reached, \tool selects the best documentation from its candidate set (Sec.~\ref{sec:selection}). It then repeats the optimization process for the next tool until all tools in the current domain have been processed.}


\subsection{Domain-Aware Execution Profiling}\label{sec:profiling}

Our empirical study shows that different task domains rely on different information fields in tool documentation. Meanwhile, a user query may require the agent to invoke multiple tools within the same domain to complete the task, making the effectiveness of one tool's documentation related to other semantically relevant tools. Therefore, we first organize tools and queries by task domain, so that subsequent documentation optimization can be performed within each domain and exploit the domain-specific information field optimization experience.

For task domain $c$, let $\mathcal{T}_c$ denote the tools in this domain~and let $\mathcal{D}^{0}_c=\{d^{0}_{t}\mid t\in\mathcal{T}_c\}$ denote the tool documentation set composed of original documentation $d^{0}_{t}$ for each tool $t$. \ly{Let $\mathcal{Q}_{c,t}$ denote the domain-associated queries whose ground-truth executions involve tool $t$. A query involving multiple tools belongs to multiple tool-specific query sets. We partition each $\mathcal{Q}_{c,t}$ into an optimization set $\mathcal{Q}^{opt}_{c,t}$ and a validation set $\mathcal{Q}^{val}_{c,t}$.} \ly{We further denote the collections of tool-specific optimization and validation query sets in domain $c$ as $\mathcal{Q}^{opt}_c= \{\mathcal{Q}^{opt}_{c,t} \mid t \in \mathcal{T}_c\}$ and $\mathcal{Q}^{val}_c= \{\mathcal{Q}^{val}_{c,t} \mid t \in \mathcal{T}_c\}$, respectively.}

Then, we execute the target LLM agent on both sets using the \ly{original tool} documentation $\mathcal{D}^{0}_c$. For queries in~$\mathcal{Q}^{opt}_{c,t}$, we collect execution traces, which include the user query, the reasoning steps,~invoked tools, tool inputs, tool outputs, and available error messages. \ly{Let $\mathcal{F}_c^0=\{\mathcal{F}_{c,t}^0 \mid t\in \mathcal{T}_c\}$ and $\mathcal{S}_c^0=\{\mathcal{S}_{c,t}^0 \mid t\in \mathcal{T}_c\}$ denote the failed and successful execution trace collections obtained using the original tool documentation $\mathcal{D}^{0}_c$, where $\mathcal{F}_{c,t}^0$ and $\mathcal{S}_{c,t}^0$ represent the failed and successful traces involving tool $t$, respectively. The failed traces are used as evidence for optimizing tool documentation, while the successful traces are retained for regression checking. For queries in validation sets, we record the initial task success rate $\text{TS}^0_c = \{ \text{TS}^0_{c,t} \mid t \in \mathcal{T}_c\}$ of the target LLM agent as a reference before documentation optimization.} Finally, the output of this stage is a profiling result $\langle \mathcal{T}_c, \mathcal{D}^{0}_c, \mathcal{Q}^{opt}_c, \mathcal{Q}^{val}_c, \mathcal{F}^0_{c}, \mathcal{S}^0_{c}, \text{TS}^0_c \rangle$ for each~task~domain.

\subsection{Memory-Guided Documentation Optimization}\label{sec:generation}

\begin{figure}[t]
    \centering
    \includegraphics[width=\linewidth]{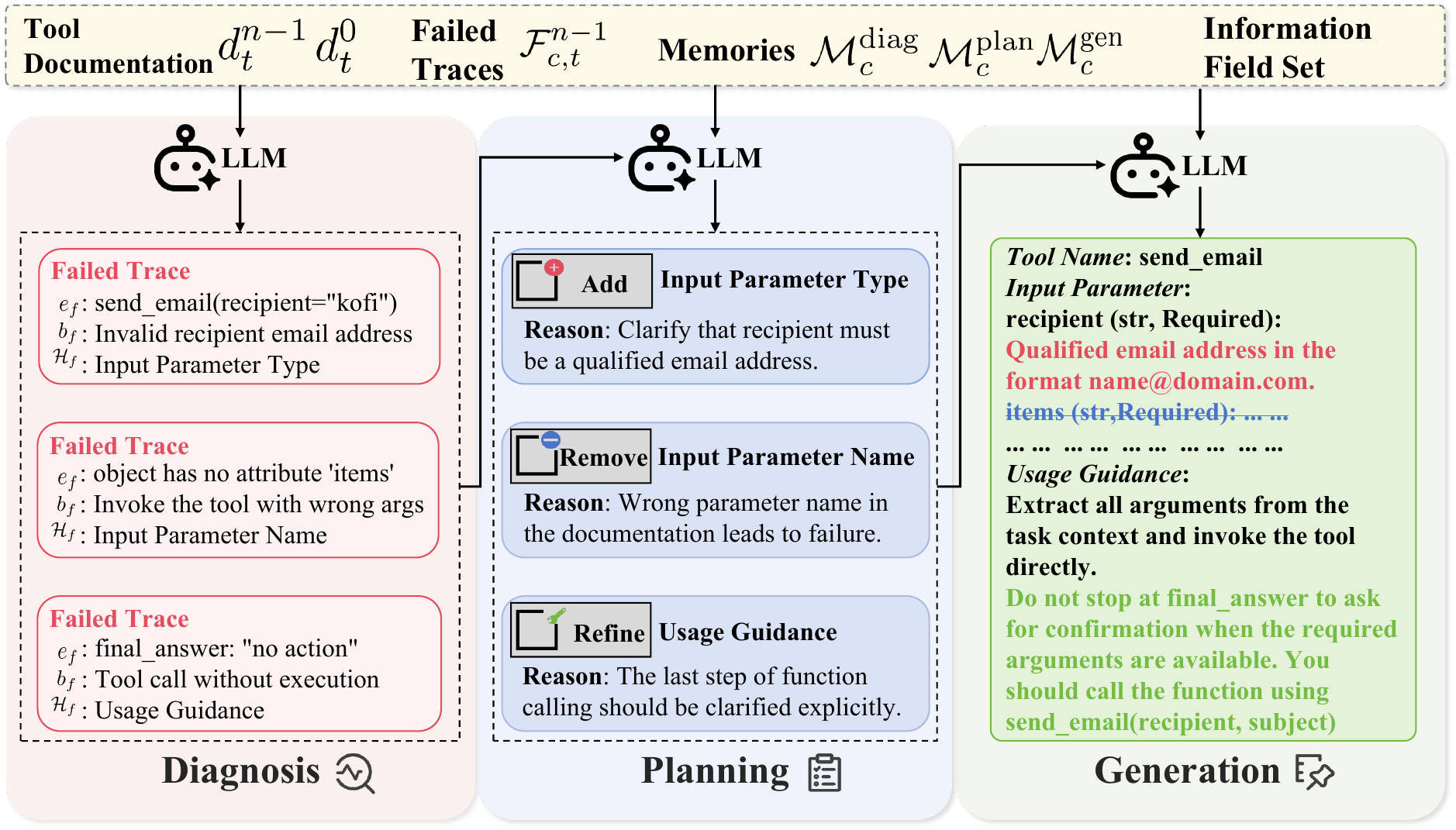}
    \caption{Memory-Guided Documentation Optimization}
    \label{fig:optimization}
\end{figure}

Within each task domain $c$, \tool optimizes the~original tool documentation $d^0_t$ of each tool $t\in\mathcal{T}_c$ iteratively. At iteration $n$, instead of directly asking an LLM to rewrite the current documentation~\cite{easytool, play2prompt}, \tool decomposes~the optimization process into three LLM-assisted steps, i.e., \textit{failed trace diagnosis}, \textit{field operation planning}, and \textit{optimized documentation generation}, as illustrated in Fig.~\ref{fig:optimization}. The diagnosis step identifies why the LLM agent fails under the current documentation. The planning step maps the diagnosed issues to possible field-level operations. The generation step applies the planned operations to produce new tool documentation~$d_t^{n}$.

\textbf{Failed Trace Diagnosis.}
At iteration $n$, \tool first diagnoses the failed execution traces \ly{$\mathcal{F}_{c,t}^{n-1}$} collected~from \ly{optimization set $\mathcal{Q}^{opt}_{c,t}$} under the latest documentation~$d_t^{n-1}$.~Since execution failures may be caused by factors~beyond tool documentation, we prompt an LLM to inspect each failed trace together with the current documentation $d_t^{n-1}$, the ground truth, and the diagnosis memory $\mathcal{M}_{c}^{\text{diag}}$ of the~current~task domain.

\begin{figure}[t]
    \centering
    \includegraphics[width=0.85\linewidth]{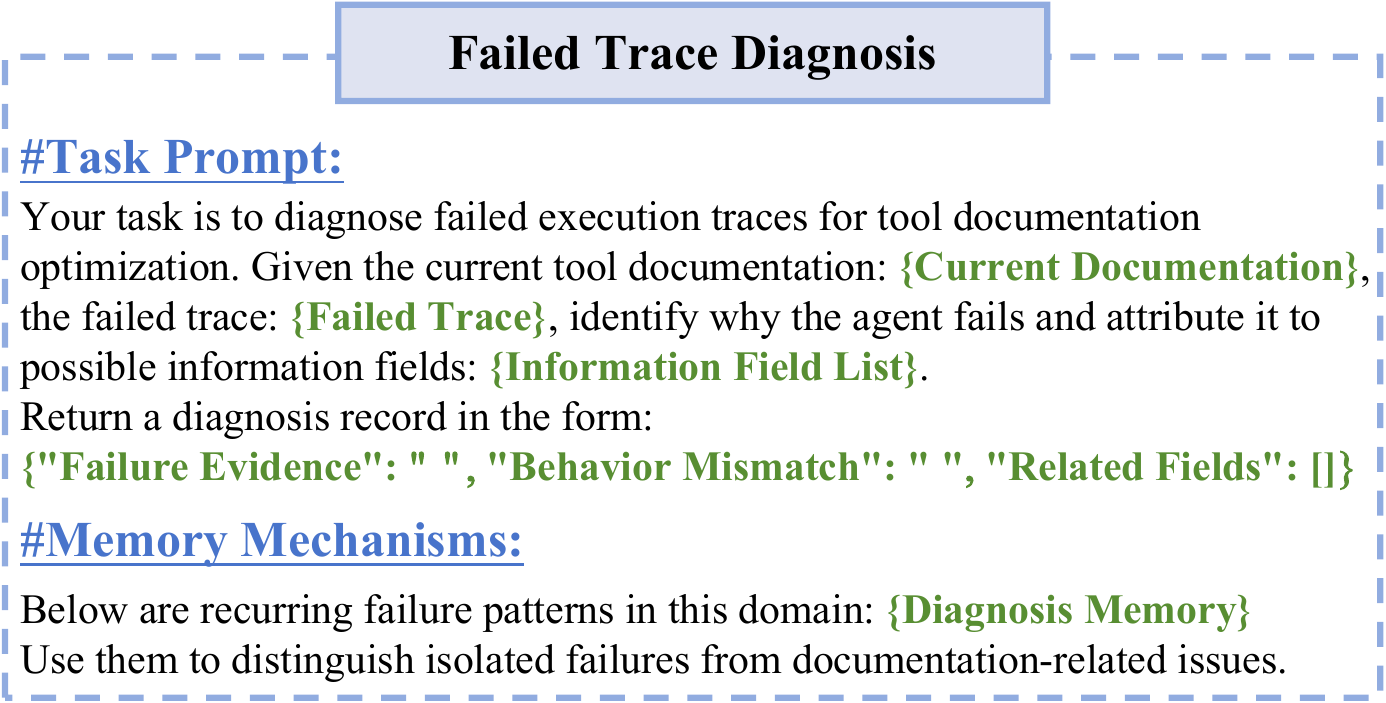}
    \caption{Prompt Template for Failed Trace Diagnosis}
    \label{fig:diagnosis}
\end{figure}

The diagnosis memory stores domain-specific failure patterns observed in previous iterations, including frequently occurring agent errors and their potentially related information fields. By referring to such memory, the LLM \ly{can reuse diagnosis experience accumulated from previous iterations, rather than analyzing each failed trace in isolation}. Fig.~\ref{fig:diagnosis} shows the template of the diagnosis prompt for the LLM. Specifically, the LLM is required to summarize the failure evidence, identify the mismatch between agent behaviors and expected tool invocations, and infer the potentially responsible information fields from those summarized in Table~\ref{tab:field_prevalence} for failure cases. 

Formally, for each failed trace \ly{$f \in \mathcal{F}_{c,t}^{n-1}$}, we obtain a diagnosis record $r_{f}=\langle e_{f}, b_{f}, \mathcal{H}_{f}\rangle$, where $e_{f}$ denotes the summarized failure evidence, $b_{f}$ denotes the agent behavior mismatch with respect to the ground truth, and $\mathcal{H}_{f}$ denotes the set of potentially related information fields. The diagnosis records of all failed traces form the set \ly{$\mathcal{R}_t^n = \{r_f\mid f \in \mathcal{F}_{c,t}^{n-1}\}$}, which is then passed to the field operation planning step.

\textbf{Field Operation Planning.}
After obtaining the diagnosis set $\mathcal{R}_t^n$, \tool further plans how to revise the tool documentation at the information-field level. Given $\mathcal{R}_t^n$, the original documentation $d_t^0$, and the planning memory $\mathcal{M}_{c}^{\text{plan}}$ of the current task domain, we prompt an LLM to map the diagnosed issues to concrete information field operations. Here, $d_t^0$ serves as the semantic anchor for planning operations, preventing the optimization process from drifting away from the original tool documentation semantics.

\begin{figure}[t]
    \centering
    \includegraphics[width=0.85\linewidth]{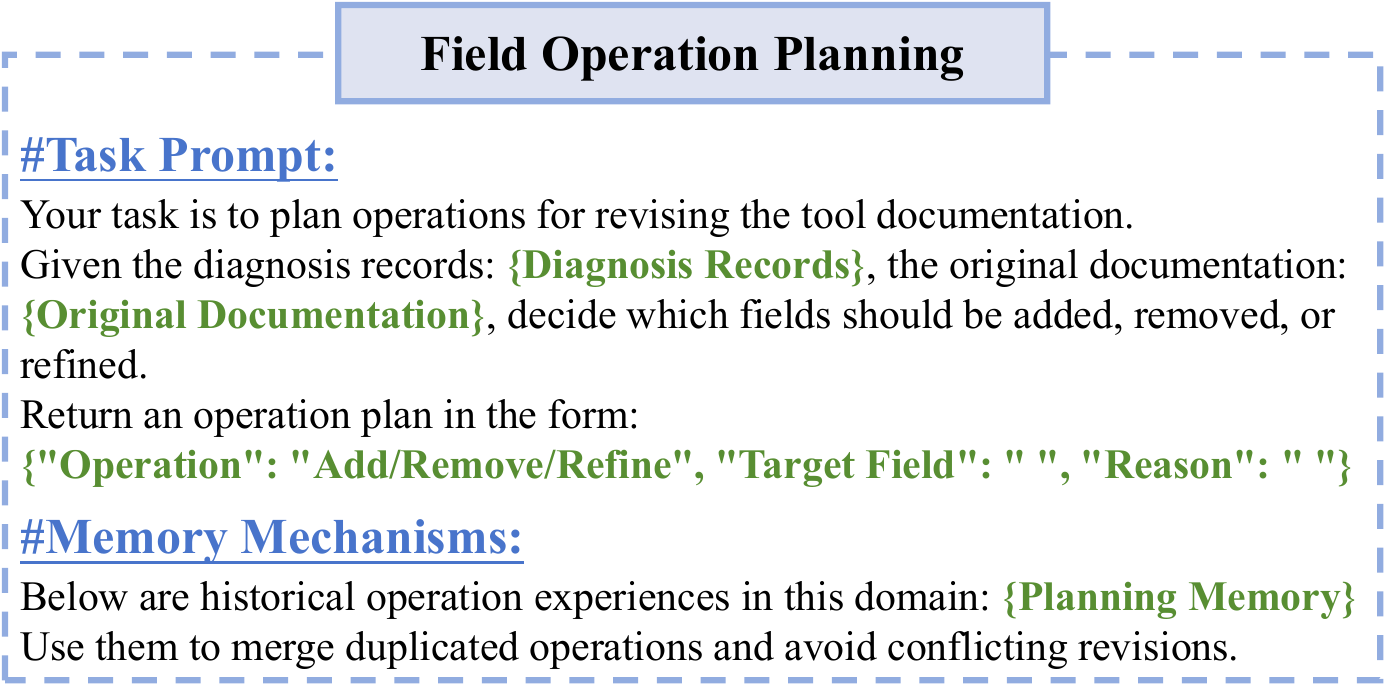}
    \caption{Prompt Template for Field Operation Planning}
    \label{fig:planning}
\end{figure}

The planning memory stores domain-specific operation experience observed in previous iterations, including which field operations are useful for addressing certain failure patterns. For example, missing invocation conditions can be mapped to adding invocation constraints, distracting usage examples or implementation snippets can be mapped to removing redundant fields, and confusion between similar tools can be mapped to refining functionality descriptions or usage guidance. By referring to such memory, \tool avoids planning field operations from each failed trace in isolation, and instead exploits historical optimization experience from related tools in the same domain. Fig.~\ref{fig:planning} shows the template of the planning prompt for the LLM. Specifically, the LLM is required to examine each diagnosis record in $\mathcal{R}_t^n$, determine whether the diagnosed issue requires adding, removing, or refining an information field, and provide the rationale for the operation.

Formally, for each diagnosis record $r_f \in \mathcal{R}_t^n$, we obtain an information-field operation plan \ly{$p_f=\{\langle o_f, h_f, g_f\rangle \mid h_f\in\mathcal{H}_f$\}}, where $o_f \in \{\textsc{Add}, \textsc{Remove}, \textsc{Refine}\}$ denotes the planned operation, $h_f \in \mathcal{H}_f$ denotes the target information field, and $g_f$ denotes the reason for applying this operation. Since different failed traces may lead to duplicated or conflicting operations, we merge trace-level plans into a unified operation plan $\mathcal{P}_t^n$ by removing duplicated operations and resolving conflicts according to their supporting diagnosis records, the planning memory, and the original documentation. The resulting plan $\mathcal{P}_t^n$ specifies which information fields should be added, removed, or refined in the next documentation~generation~step.

\textbf{Optimized Documentation Generation.}
Given the unified operation plan $\mathcal{P}_t^n$, \tool generates new tool documentation for tool $t$, using the original documentation $d_t^0$ as the generation anchor to preserve the original tool semantics and reduce the risk of information drift during repeated~rewriting.

\begin{figure}[t]
    \centering
    \includegraphics[width=0.85\linewidth]{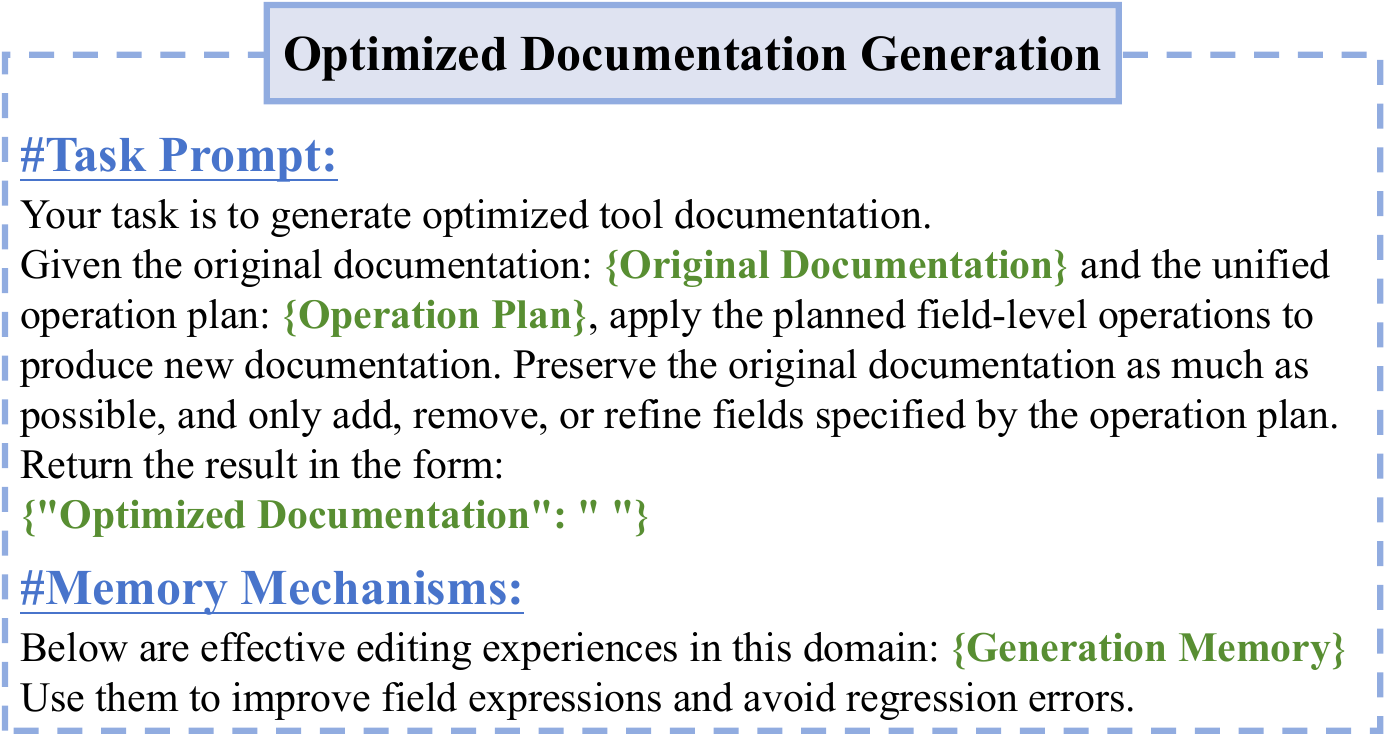}
    \caption{Prompt Template for Documentation Generation}
    \label{fig:generate}
\end{figure}

To guide the generation process, \tool further~incorporates the generation memory $\mathcal{M}_{c}^{\text{gen}}$ of the current task domain. The generation memory stores domain-specific editing experience observed in previous iterations, including \ly{field expressions that are effective for mitigating certain failure patterns}, generation constraints, and edits that may introduce regression errors. Fig.~\ref{fig:generate} shows the template of the generation prompt for the LLM.
Specifically, the LLM is required to apply the operations in $\mathcal{P}_t^n$ to $d_t^0$ by adding missing fields, removing distracting or redundant fields, and refining ambiguous or incomplete fields, while preserving unchanged information that is not targeted by the operation~plan. 

\subsection{Candidate Evaluation and Feedback Update}\label{sec:feedback}
\label{sec:iterative_refinement}

\ly{After generating the candidate documentation $d_t^n$, we~evaluate whether it improves the task success rate of the target LLM agent through agent re-execution, and update the optimization feedback for subsequent iterations. Since the optimized documentation may improve some failed queries while harming previously successful ones, we consider both validation performance and regression risk during the evaluation.}

\textbf{Regression-Aware Evaluation.} 
For the target tool $t$, we replace its original documentation $d_t^0$ with the new documentation $d_t^n$ while keeping the documentation of other tools unchanged. Then, we execute the LLM agent on~\ly{$\mathcal{Q}^{val}_{c,t}$} and compute the task success $\text{TS}_{c,t}^n$. In addition, we check whether $d_t^n$ introduces regression errors on previously successful optimization queries by re-executing the LLM agent on~\ly{$\mathcal{Q}^{opt}_{c,t}$} to obtain the failed trace set \ly{$\mathcal{F}_{c,t}^n$} and the successful trace set \ly{$\mathcal{S}_{c,t}^n$}. Based on these results, we obtain the regression set \ly{$\mathcal{G}_{c,t}^n$}, which contains the failed traces of queries that were successfully completed under the original documentation $d_t^{0}$ but failed under $d_t^n$.

\textbf{Candidate and Memory Update.}
We compare the newly generated documentation $d_t^n$ with the original documentation $d_t^0$ using the validation task success rate \ly{$\text{TS}_{c,t}^0$} obtained during execution profiling. Only when \ly{$\text{TS}_{c,t}^n \ge  \text{TS}_{c,t}^0$}, we add the new tool documentation $d_t^n$ to the candidate documentation set $\mathcal{D}^{t}_c$. 

\begin{figure}[t]
    \centering
    \includegraphics[width=0.85\linewidth]{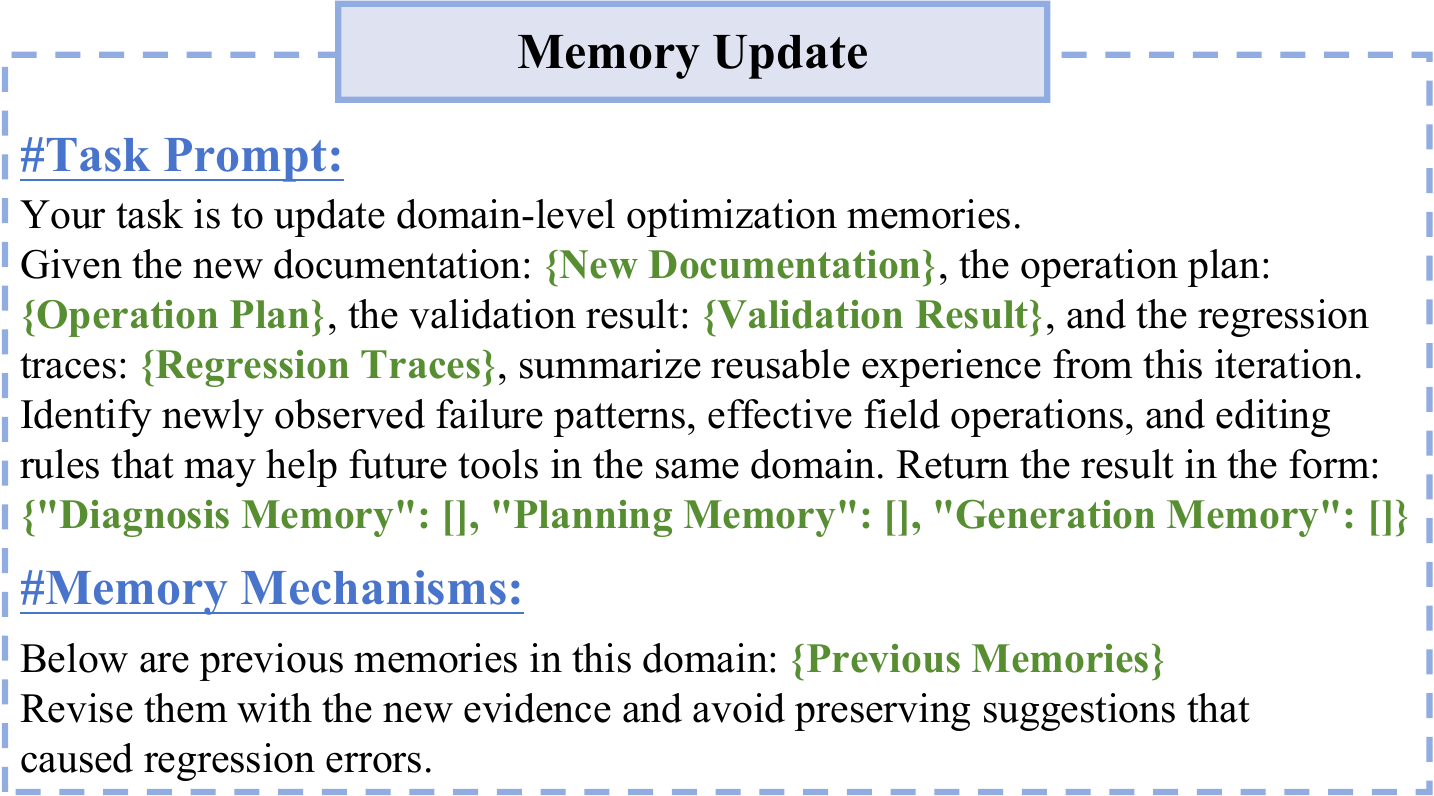}
    \caption{Prompt Template for Updating the Memory}
    \label{fig:memory_update}
\end{figure}

Meanwhile, the failed traces \ly{$\mathcal{F}_{c,t}^n$} collected on \ly{$\mathcal{Q}^{opt}_{c,t}$} are used as the input for failed trace diagnosis in the next iteration. We further update the domain-level memory based on the evaluation results. Specifically, we prompt an LLM to analyze the new documentation $d_t^n$, the regression set \ly{$\mathcal{G}_{c,t}^n$} and the validation result \ly{$\text{TS}_{c,t}^n$}. Fig.~\ref{fig:memory_update} shows the prompt template for the LLM. Specifically, we require the LLM to summarize the newly observed failure patterns, effective field operations and editing experience during this optimization iteration. Then, we update the $\mathcal{M}^{\text{diag}}_c$, $\mathcal{M}^{\text{plan}}_c$ and $\mathcal{M}^{\text{gen}}_c$, respectively. In this way, the updated traces and memories provide feedback for the next iteration, forming a closed optimization loop.

\subsection{Final Documentation Selection}\label{sec:selection}

When optimization for tool $t$ reaches the maximum number of iterations, we select the candidate with the highest validation task success rate from $\mathcal{D}_c^t$, breaking ties by choosing the~shortest documentation to reduce execution context cost. We~then~proceed to the next tool until all tools in $\mathcal{T}_c$~are~processed.

\section{Evaluation}\label{sec:evaluation}
We implement a prototype of \tool with \todo{2,687} lines of Python code. To evaluate the effectiveness and efficiency of \tool, we design the following research questions.

\begin{itemize}[leftmargin=*]
    \item \textbf{RQ3 Effectiveness Evaluation.} What is the effectiveness~of \tool in improving task success rate of LLM agents?
    \item \textbf{RQ4 Efficiency Evaluation.} What is the efficiency and optimization overhead of \tool?
    \item \textbf{RQ5 Sensitivity Analysis.} How do optimization models and iteration budgets affect the effectiveness of \tool?
    \item \textbf{RQ6 Ablation Study.} What is the contribution of our memory mechanism to the effectiveness of \tool?
\end{itemize}

\subsection{Evaluation Setup}\label{sec:setup}

\textbf{Dataset.} We select two tool-use benchmarks, \ie WorkBench~\cite{workbench} and API-Bank~\cite{apibank}. In addition to the five task domains from WorkBench used in \textbf{RQ2}, API-Bank provides tools from four additional domains, \ie finance management, health management, information retrieval, and smart home management (SHM). Overall, our evaluation dataset covers 9 task domains, involving \ly{74} tools and \ly{2,072} user queries. We split the user queries into optimization, validation, and test sets with a ratio of 5:1:4. The optimization and validation sets are used during documentation optimization, while the test set is held out throughout the optimization process and used only for the final evaluation of the optimized tool documentation. To ensure a fair and representative evaluation, we perform a stratified split over task domains and tools, so that each subset preserves the overall distribution of tool-use scenarios as much as possible. After splitting, the optimization, validation, and test sets contain \ly{1,036}, \ly{207}, and \ly{829} user queries, respectively. The detailed statistics of the evaluation dataset across task domains and tools are available at~\cite{website} due to space limitations.

\textbf{Metrics.} 
Beyond the $\text{TS}$ metric used in our empirical study, we also use \textit{tool invocation correctness}~($\text{TC}$) to measure whether the~LLM agent invokes the correct tools. Specifically, given the test suite containing tool-use queries $\mathcal{Q}$, for each query ${q_i} \in \mathcal{Q}$, we denote the ground-truth tool set as $T_{q_i}$ and the tool set predicted by the LLM agent as $\hat{T}_{q_i}$. A query is regarded as tool-correct only when the predicted tool set exactly matches the ground-truth set, \ie $T_{q_i} = \hat{T}_{q_i}$. We compute tool invocation correctness as $\text{TC} = \frac{1}{|\mathcal{Q}|}\sum_{i=1}^{|\mathcal{Q}|}\mathbb{I}(T_{q_i} = \hat{T}_{q_i})$, where $\mathbb{I}(\cdot)$ is the indicator function. Together, $\text{TC}$ and $\text{TS}$ capture the effectiveness of tool documentation on both tool invocation behavior and task completion performance of~LLM~agents.

\textbf{Baseline.} We compare \tool with two tool documentation optimization approaches, \ie \textsc{EasyTool}~\cite{easytool} and \textsc{DRAFT}~\cite{qu2025exploration}. \ly{\textsc{EasyTool} uses an LLM and a predefined template to convert documentation into concise, structured tool-use instructions. \textsc{DRAFT} uses LLMs to iteratively refine tool documentation based on feedback obtained by generating tool inputs and executing the tools}. We use their official artifacts and adapt only the input interfaces to our dataset.

\textbf{RQ Setup.} 
For \textbf{RQ3} and \textbf{RQ4}, we apply \tool with Claude Haiku 4.5 as the optimization model to optimize the original tool documentation in our dataset. \ly{We set the maximum number of iterations to 5 by default to balance optimization effectiveness and LLM token overhead.} For \textbf{RQ3}, we evaluate the effectiveness of \tool by comparing it with using the original tool documentation and using the tool documentation optimized by \textsc{EasyTool} and \textsc{DRAFT}, respectively. To understand whether the improvement generalizes across different agent settings, we conduct comparisons from three perspectives introduced in \textbf{RQ2}, \ie task domains, LLM backbones, and agent paradigms, respectively.

For \textbf{RQ4}, we report the additional token cost introduced by the optimized documentation during agent execution, compared with the original documentation. We also measure the optimization overhead of \tool, including the time cost and token consumption required to generate optimized tool documentation, and compare them with the baselines.

For \textbf{RQ5}, we examine how different optimization models and iteration budgets affect \tool in the data analysis domain. We vary the LLM used for documentation optimization, \ie GPT-4o and GLM-5, and the maximum number of optimization iterations, while keeping the original tool documentation, query set, and target agent unchanged. The target agent is a GPT-4o-based ReAct agent. We report the average $\text{TC}$ and $\text{TS}$ on the test set using the optimized documentation generated under each setting.

For \textbf{RQ6}, we conduct an ablation study to evaluate the contribution of our memory mechanism. We~construct a variant of \tool, denoted as \tool-NoMem, by disabling the memory mechanism while keeping the remaining optimization pipeline unchanged. We then compare the tool documentation in the data analysis domain optimized by \tool and \tool-NoMem, and report the average $\text{TC}$ and $\text{TS}$ of the GPT-4o-based ReAct agent, analyzing whether the memory mechanism helps \tool produce more effective tool documentation for the target LLM agent.

\textbf{Environment.} We conduct all the experiments on Ubuntu 20.04.4 LTS servers with 4 NVIDIA GeForce RTX 3090 GPUs, Intel(R) Xeon(R) Silver 4310 @ 2.10GHz and 128GB memory.

\begin{figure*}[t]
    \captionsetup[subfigure]{font=scriptsize}
    \centering
    \subfloat[Data Analysis+GPT+ReAct]{
        \includegraphics[width=0.19\textwidth]{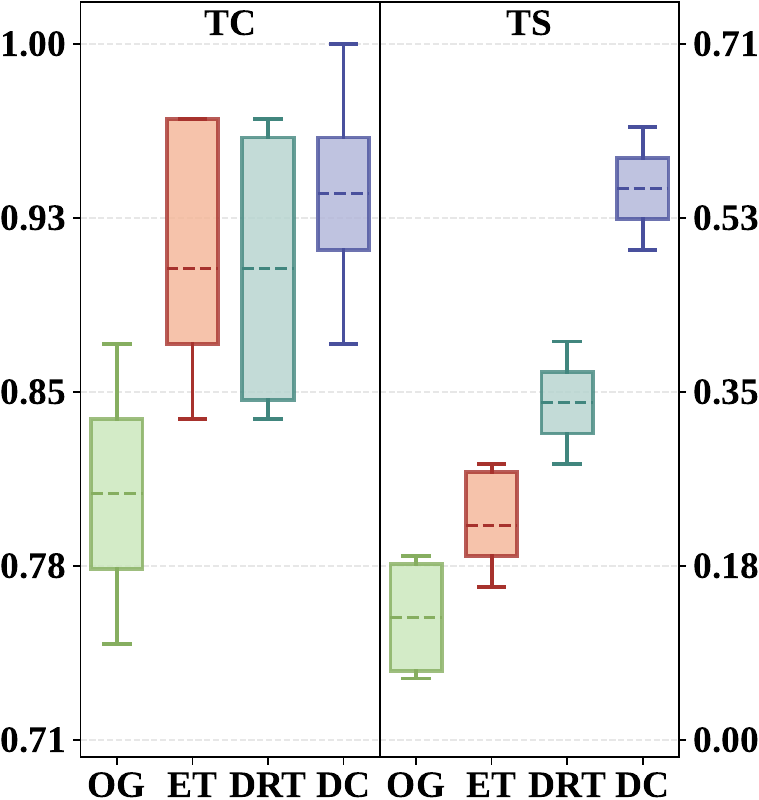}
        \label{img:rq3-data}
    }
    \subfloat[Email+GPT+ReAct]{
        \includegraphics[width=0.19\textwidth]{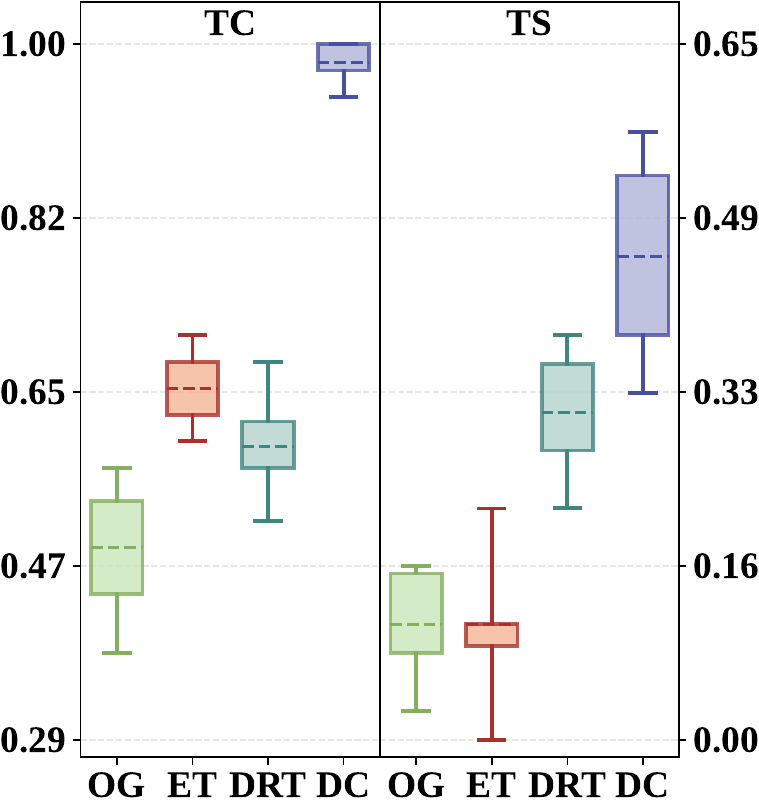}
        \label{img:rq3-email}
    }
    \subfloat[Calendar+GPT+ReAct]{
        \includegraphics[width=0.19\textwidth]{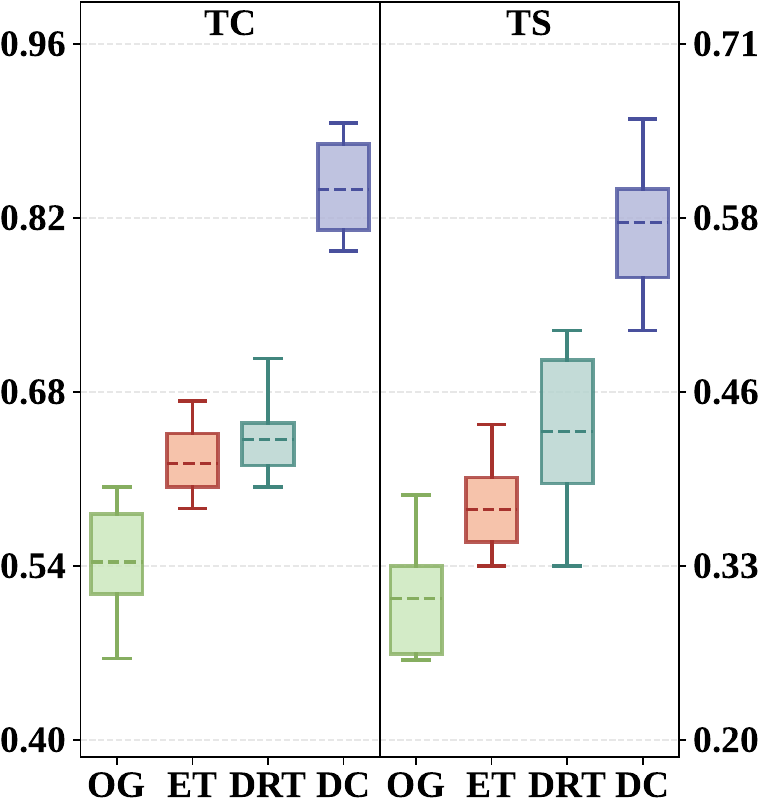}
        \label{img:rq3-calendar}
    }
    \subfloat[CRM+GPT+ReAct]{
        \includegraphics[width=0.19\textwidth]{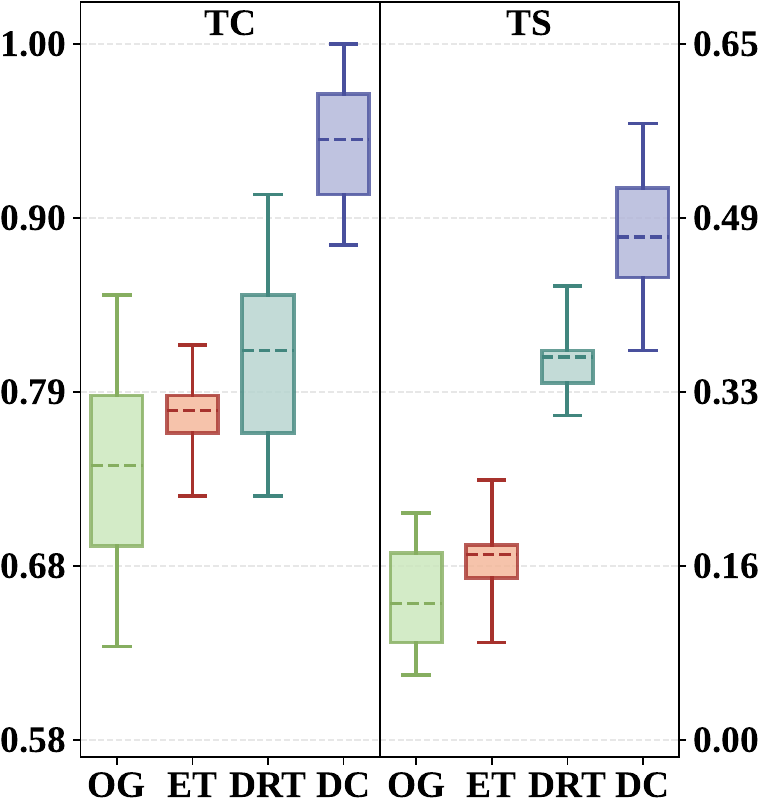}
        \label{img:rq3-crm}
    }
    \subfloat[Health+GPT+ReAct]{
        \includegraphics[width=0.19\textwidth]{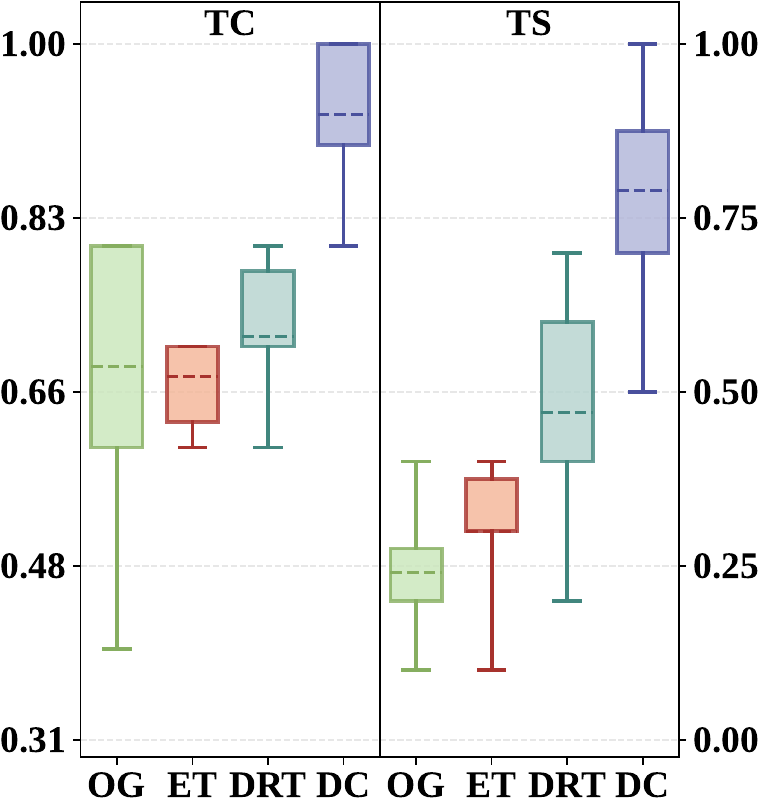}
        \label{img:rq3-health}
    }

    \subfloat[Information+GPT+ReAct]{
        \includegraphics[width=0.19\textwidth]{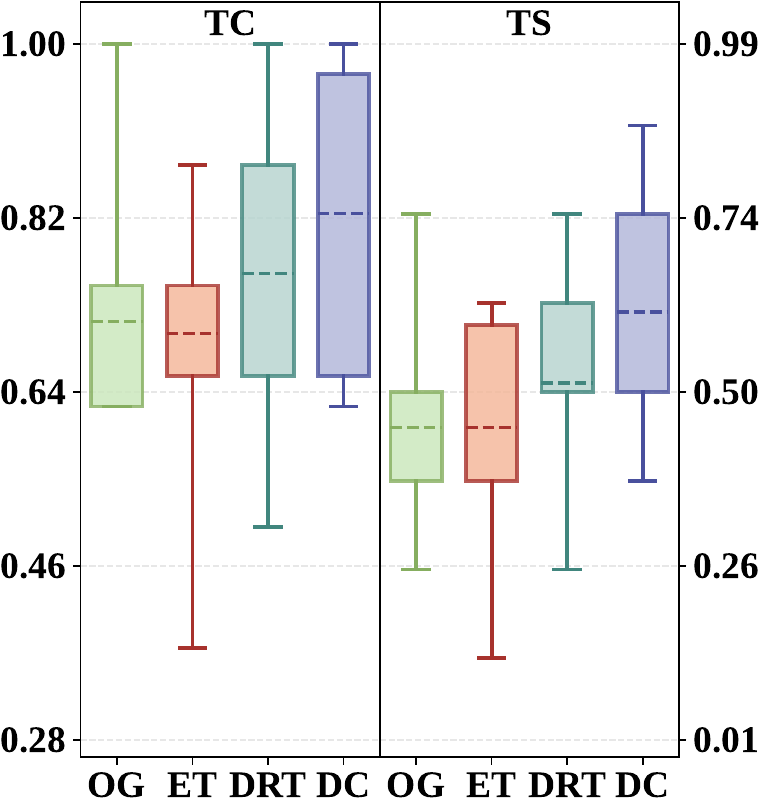}
        \label{img:rq3-information}
    }
    \subfloat[SHM+GPT+ReAct]{
        \includegraphics[width=0.19\textwidth]{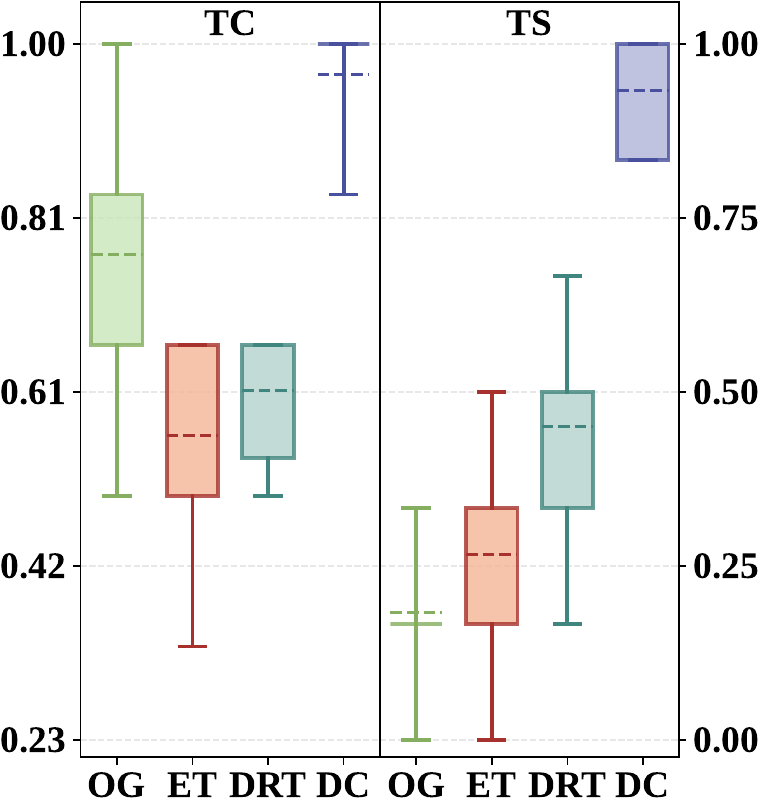}
        \label{img:rq3-shm}
    }
    \subfloat[Data Analysis+GLM+ReAct]{
        \includegraphics[width=0.19\textwidth]{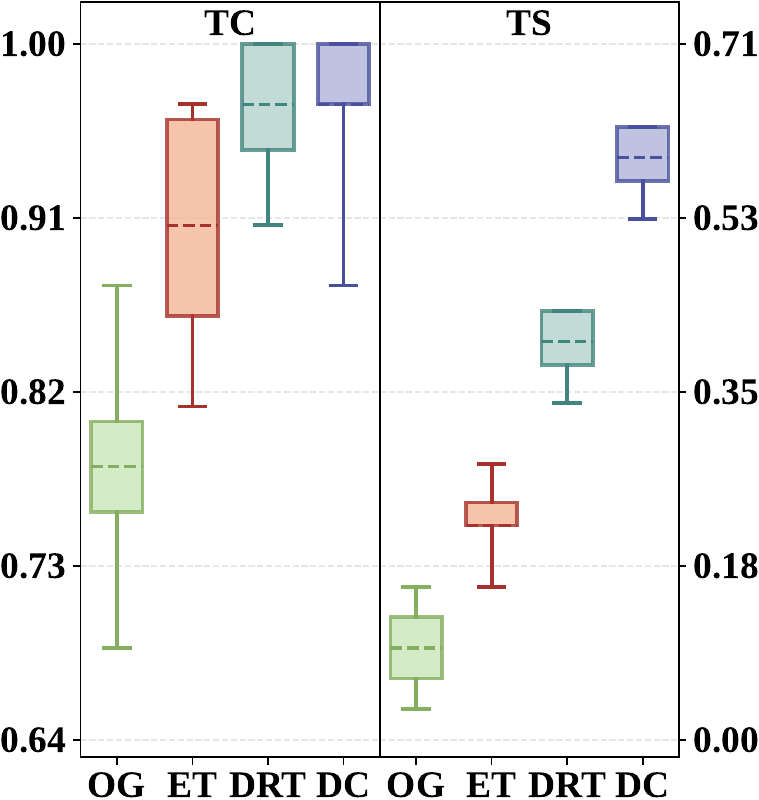}
        \label{img:rq3-glm}
    }
    \subfloat[Data Analysis+Claude+ReAct]{
        \includegraphics[width=0.19\textwidth]{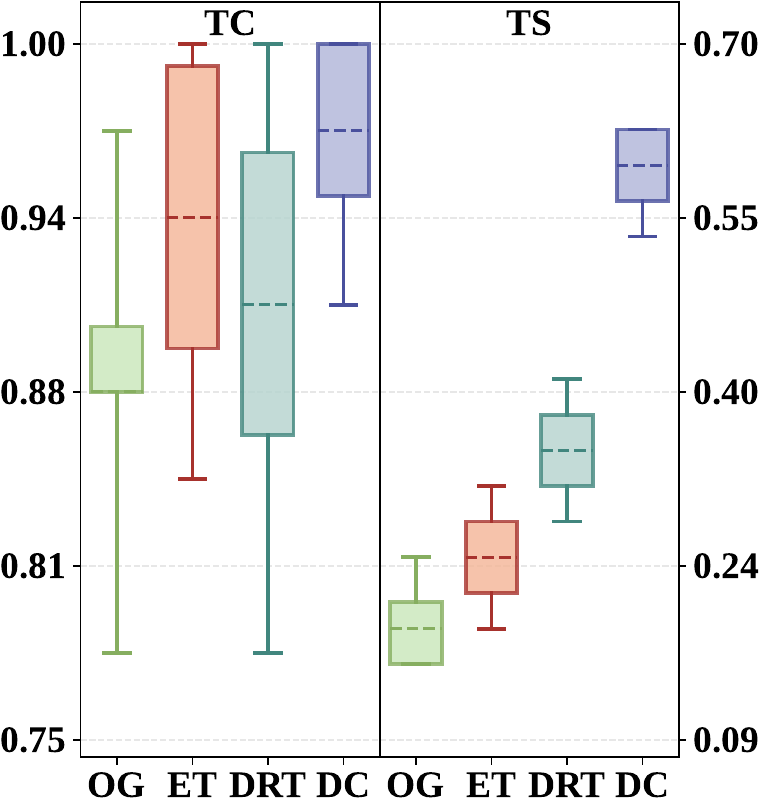}
        \label{img:rq3-claude}
    }
        \subfloat[Data Analysis+GPT+Multi-Agent]{
        \includegraphics[width=0.19\textwidth]{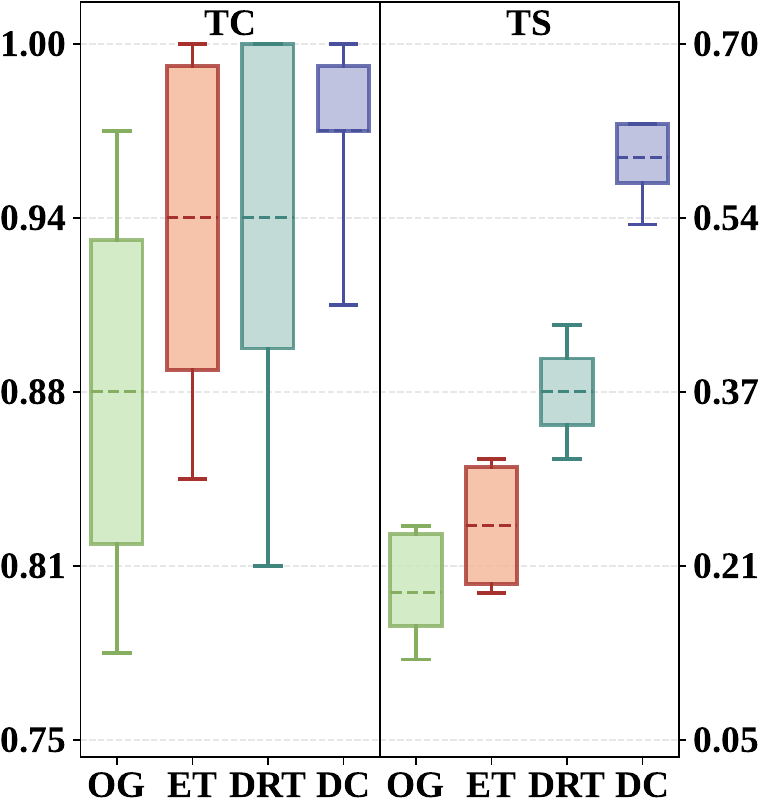}
        \label{img:rq3-multiagent}
    }    
\caption{Effectiveness Comparison across Task Domains, LLM Backbones, and Agent Paradigms}
\label{fig:effectiveness}
\end{figure*}

\subsection{Effectiveness Evaluation (RQ3)}\label{sec:rq3}

\textbf{Overall Results.}
Fig.~\ref{fig:effectiveness} compares \tool (DC) with the original documentation (OG), \textsc{EasyTool} (ET), and \textsc{DRAFT} (DRT) across task domains, LLM backbones, and agent paradigms. We independently repeat the optimization process three times. For each optimized result, we conduct five independent evaluation runs \todo{on the test set} to account~for~execution randomness. Hence, each boxplot summarizes the results from $3 \times 5 = 15$ runs. Overall, \tool achieves the highest average \text{TC} and \text{TS} in all evaluated settings. Compared with the original documentation, \tool improves \text{TC} by \todo{34.69\%} and \text{TS} by \todo{95.89\%} on average. Compared with the two baseline approaches, it achieves an average \text{TC} improvement of \todo{30.83\%} and \text{TS} improvement of \todo{75.15\%}. Even against the strongest baseline in each setting, \tool improves \text{TC} and \text{TS} by \todo{26.23\%} and \todo{46.11\%} on average, respectively.

With respect to task domains, Fig.~\ref{img:rq3-data}-\ref{img:rq3-shm} show that \tool consistently achieves the highest average \text{TC} and \text{TS} in the seven domains presented in the paper. The remaining two domains, \ie project management and finance management, are reported on our website~\cite{website} due to space limitations. Across all nine evaluated domains, \tool improves \text{TC} and \text{TS} by \todo{46.27\%} and \todo{194.22\%} over the original documentation, respectively, and by \todo{27.34\%} and \todo{42.36\%} over the strongest documentation optimization baseline on average. The gains are particularly evident in domains such as email management, customer relationship management, health management, and smart home management, where the original and baseline documentation often yields low task success rate despite~moderate tool~invocation~correctness. 

With respect to LLM backbones, Fig.~\ref{img:rq3-data}, Fig.~\ref{img:rq3-glm}, and Fig.~\ref{img:rq3-claude} show that \tool remains best-performing for all three backbones.~Compared with the strongest baseline, it improves \text{TC} by \todo{3.45\%} and \text{TS} by \todo{60.83\%}, on average, across GPT-4o, GLM-5, and Claude Haiku 4.5.

With respect to agent paradigms, Fig.~\ref{img:rq3-data} and Fig.~\ref{img:rq3-multiagent} compare ReAct and Multi-Agent under the same task domain and GPT-4o backbone. \tool achieves the highest \text{TC} and \text{TS} under both paradigms. Compared with the strongest baseline, it improves \text{TC} by \todo{3.44\%} and \todo{3.34\%}, and \text{TS} by \todo{63.61\%} and \todo{58.35\%} under ReAct and Multi-Agent, respectively.

The results further show that \tool generally achieves higher lower-end performance and more compact distributions across repeated runs. Compared with \textsc{EasyTool} and \textsc{DRAFT}, \tool reduces the average interquartile range of \text{TC} by \todo{41.56\%} and \todo{52.93\%}, respectively, and that of \text{TS} by \todo{36.71\%} and \todo{44.07\%}. These results indicate that \tool introduces stable improvements across different~agent~settings.

\textbf{Breakdown Analysis.} 
\textsc{EasyTool} improves some settings by transforming documentation into concise and structured instructions. However, its fixed template applies the same information fields across agent settings. It may remove fields useful to a particular agent or fail to introduce fields absent from the original documentation, resulting in limited improvements.

\textsc{DRAFT} uses feedback obtained by generating tool inputs and executing tools to iteratively refine tool documentation. Although such feedback helps correct tool descriptions, it mainly reflects isolated tool executions rather than failures occurring during end-to-end agent task solving. Moreover, it does not explicitly adapt the composition of information fields. Consequently, documentation-related failures observed in complete agent traces may remain unresolved.

In contrast, \tool diagnoses failed agent execution traces and performs field-level addition, removal, and refinement according to the target agent setting, allowing for better alignment of tool documentation with the information requirements of the target agent, explaining its consistent improvements in both \text{TC} and \text{TS}. Nevertheless, \tool does not achieve perfect results. Some failures originate from factors beyond tool documentation, such as incorrect task decomposition, stochastic LLM reasoning, or incomplete user queries. Therefore, documentation optimization can substantially improve agent effectiveness, but cannot eliminate failures caused by other components of the agent~execution~process.

    \textit{\textbf{Summary.}} \tool consistently outperforms \textsc{EasyTool} and \textsc{DRAFT} across task domains, LLM backbones, and agent paradigms, improving tool invocation correctness by \todo{30.83\%} and task success rate by \todo{75.15\%}, on average.

\subsection{Efficiency Evaluation (RQ4)}\label{sec:rq4}

\begin{table}[t]
    \centering
    \small
    \caption{Results of Efficiency Evaluation}
    \begin{tabular}{lccc}
    \toprule
    \textbf{Approach} &
    \textbf{Doc Length} &
    \textbf{Token Cost} &
    \textbf{Time Cost (min)} \\
    \midrule
    Original   & 152.58 & -- & -- \\
    \textsc{EasyTool}   & 82.68 & 237.81 & 0.23 \\
    \textsc{DRAFT}      & 189.56 & 4,784.39 & 2.26 \\
    \tool & 189.23 & 4,480.81 & 12.65 \\
    \bottomrule
    \end{tabular}
    \label{tab:efficiency_table}
\end{table}

Table~\ref{tab:efficiency_table} reports the average documentation length and optimization overhead per tool. \textsc{EasyTool} produces the shortest documentation because it primarily compresses the original content. In contrast, \textsc{DRAFT} and \tool generate slightly longer documentation by incorporating additional information. The documentation produced by \tool contains 189.23 tokens on average, which is \todo{24.02\%} longer than the original documentation but comparable to \textsc{DRAFT}. Thus, \tool introduces only a limited additional context cost.

For optimization overhead, \tool consumes \todo{6.35\%} fewer tokens than \textsc{DRAFT} for each tool during optimization. However, it requires 12.65 minutes on average, longer than \textsc{EasyTool} and \textsc{DRAFT}, because \tool repeatedly diagnoses failed traces, generates candidate documentation, and evaluates candidates through agent re-execution. Overall, \tool trades additional offline optimization time for substantially improved effectiveness, while maintaining comparable documentation length and token consumption to~\textsc{DRAFT}.

    \textit{\textbf{Summary.}} \tool incurs a higher optimization-time cost, requiring 12.65 minutes per tool on average, while consuming \todo{6.35\%} fewer optimization tokens than DRAFT and producing documentation of comparable length.

\subsection{Sensitivity Analysis (RQ5)}\label{sec:RQ5}

\begin{table}[t]
    \centering
    \small
    \caption{Results of Model Sensitivity}
    \begin{tabular}{lccc}
    \toprule
    \textbf{Metrics} &
    \textbf{GPT-4o} &
    \textbf{GLM-5} &
    \textbf{Claude Haiku 4.5} \\
    \midrule
    TC   & 47.50\% & 73.75\% & 93.75\% \\
    TS   & 21.25\% & 35.63\% & 56.25\%\\ 
    \bottomrule
    \end{tabular}
    \label{tab:model_sensitivity}
\end{table}

\begin{figure}[t]
    \centering
    \includegraphics[width=0.8\linewidth]{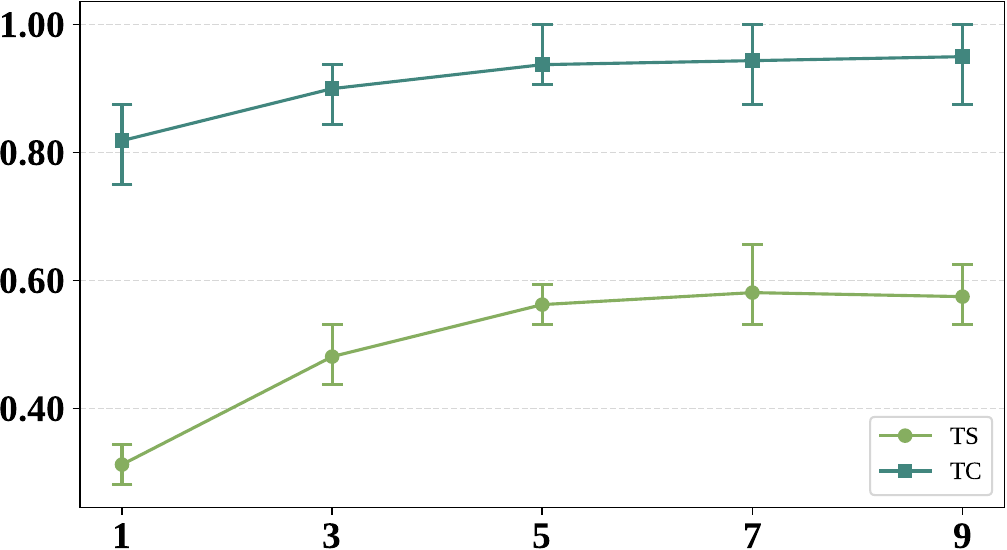}
    \caption{Results of Iteration Sensitivity}
    \label{fig:params_sensitivity}
\end{figure}

\textbf{Model Sensitivity.}
Table~\ref{tab:model_sensitivity} reports the effectiveness of \tool using different optimization models while keeping the target agent and evaluation setting unchanged. Claude Haiku 4.5 achieves the best performance, improving \text{TC} and \text{TS} over GLM-5 by \todo{27.12\%} and \todo{57.87\%}, respectively, and over GPT-4o by \todo{97.37\%} and \todo{164.71\%}. GLM-5 also outperforms GPT-4o, with relative gains of \todo{55.26\%} in \text{TC} and \todo{67.67\%} in \text{TS}. These results indicate that the quality of optimized documentation is sensitive to the optimization model. Documentation optimization requires the model to diagnose failed traces, identify field-level deficiencies, and generate appropriate revisions. Models with stronger instruction-following and reasoning capabilities can produce more effective~documentation.

\textbf{Iteration Sensitivity.}
Fig.~\ref{fig:params_sensitivity} shows that increasing the number of optimization iterations improves both \text{TC} and \text{TS}. The gains are most pronounced in the early rounds. Compared with one iteration, five iterations improve \text{TS} by \todo{80.00\%} and \text{TC} by \todo{14.50\%}. This indicates that iterative execution feedback progressively helps \tool identify and address documentation deficiencies. After five iterations, however, the marginal gains become small, suggesting that the optimization process has largely converged. We therefore set the default iteration budget to five to avoid large optimization~overhead.

\textit{\textbf{Summary.}} Stronger optimization models generally yield better documentation quality, while iterative refinement substantially improves optimization effectiveness. The budget of five iterations achieves a favorable trade-off.

\subsection{Ablation Study (RQ6)}\label{sec:rq6}

\begin{figure}[t]
    \centering
    \includegraphics[width=0.8\linewidth]{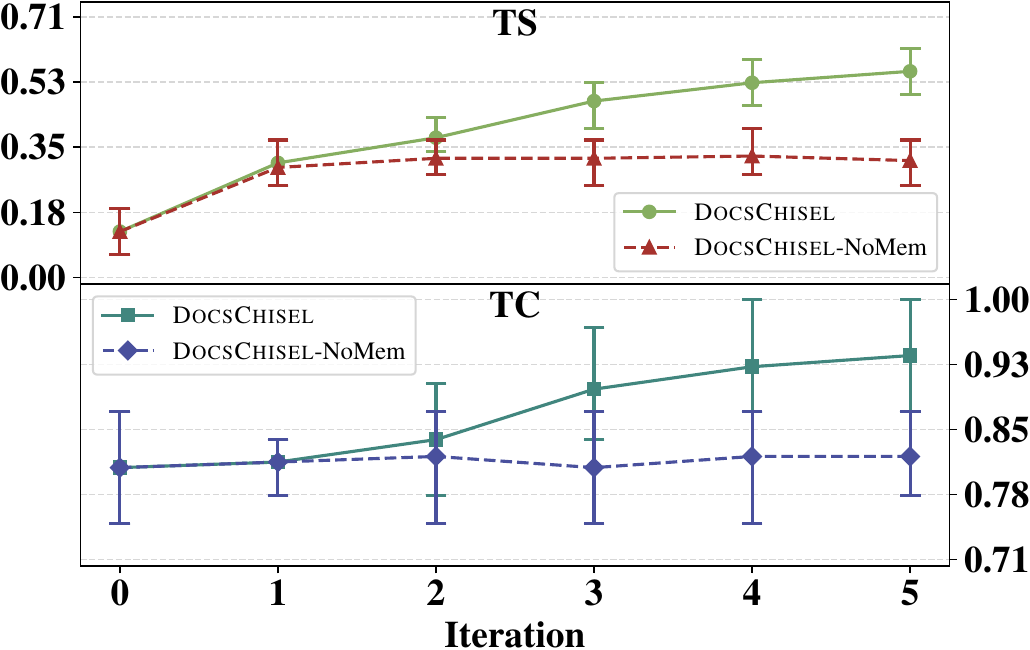}
    \caption{Results of Ablation Study}
    \label{fig:memory_ablation}
\end{figure}

Fig.~\ref{fig:memory_ablation} compares \tool with \tool-NoMem over five optimization iterations. In the first two iterations, the two variants achieve similar performance, indicating that both can improve documentation using the feedback from the current iteration. However, their performance diverges as optimization proceeds. At the fifth iteration, \tool-NoMem reduces \text{TC} and \text{TS} by \todo{13.64\%} and \todo{76.47\%} compared with \tool, respectively. Moreover, the relative \text{TS} improvement between \tool and \tool-NoMem increases from \todo{4.17\%} at Iteration-1 to \todo{76.47\%} at Iteration-5, indicating that the benefit of memory becomes more pronounced as optimization experience accumulates. Without memory, both metrics largely plateau after the second iteration. In contrast, \tool continues to improve by reusing previously observed documentation deficiencies, effective field operations, and editing experience. Such experience guides subsequent diagnosis, field planning, and candidate generation, preventing the optimizer from repeatedly exploring ineffective~revisions.

    \textit{\textbf{Summary.}} The memory mechanism is critical to iterative documentation optimization, improving the final \text{TC} and \text{TS}, on average, by \todo{13.64\%} and \todo{76.47\%}, respectively.


\section{Threats to Validity}\label{sec:threats}

First, the selection of agent paradigms, LLM backbones, and datasets may limit the generalizability of our findings. To mitigate this threat, we evaluate \tool using two agent paradigms with three LLM backbones across nine task~domains.

Second, our evaluation metrics may not capture all aspects of agent tool-use effectiveness. To reduce this threat, we report complementary metrics that evaluate both tool invocation correctness and final task success rate. Moreover, identifying information fields and constructing documentation variants involve manual inspection and correction, which may introduce subjective bias. To mitigate this threat, multiple authors independently conducted the analysis and resolved disagreements through discussion, achieving a Cohen's kappa coefficient of 0.862, with strong inter-rater agreement.

Finally, the inherent randomness of LLMs and the instability of agent execution pose additional threats to validity. To mitigate these threats, we keep the prompts and execution environments unchanged within each comparison, use the official replication artifacts of all baselines, and conduct three independent optimization runs, each followed by five evaluation runs. Thus, each evaluated setting is assessed over $3 \times 5 = 15$ runs. Across our evaluation, we apply the Mann-Whitney U~test~\cite{mcknight2010mann}, and the improvements of \tool remain statistically significant after Holm correction~\cite{holm1979simple}~($\mathrm{p}_{\mathrm{adj}} < 0.05$).

\section{Related Work}\label{sec:related-work}

\subsection{LLM Agent Ecosystems}
Recent advances in LLM agents and agent ecosystems have substantially improved the capability of LLMs to interact with external tools~\cite{mavroudis2024langchain, dify, hong2024metagpt, he2025llm, xia2025demystifying, claudehaiku}. Specifically, ReAct~\cite{yao2022react} introduces the reasoning-and-acting paradigm for iterative tool invocation, while Toolformer~\cite{schick2023toolformer} demonstrates that LLMs can learn tool-use behaviors through self-supervised training. AutoGen~\cite{wu2024autogen} provides a unified framework for multi-agent orchestration and workflow automation. Subsequent works~\cite{apibench, apibank, toolllm} further improve large-scale tool-use capabilities of LLM agents. Representative agent systems such as Claude Code~\cite{anthropic2025claudecode}, OpenAI Codex~\cite{openai2025codex}, and SWE-agent~\cite{yang2024swe} have demonstrated the growing applicability of tool-use LLM agents in software engineering and other complex domains~\cite{lin2025soen}. To evaluate the reliability and execution capability of tool-use LLM agents, recent studies~\cite{toolbench, taubench, workbench, anytoolbench, shortcutsbench, yu2026wildtoolbench, jimenez2024swe, crmarena, ToolAlpaca, toole} propose diverse benchmarks that include tools with documentation.

Although these studies significantly advance tool-use capabilities of LLM agents, they mainly focus on tool construction, retrieval, invocation, and evaluation strategies, while largely treating tool documentation as fixed input for tool grounding and execution. In contrast, our work systematically investigates the structure and effectiveness of tool documentation itself, and studies how adaptive tool documentation optimization affects downstream task success rate across different agent settings.

\subsection{Tool Documentation Engineering}
Software engineering research has long recognized the importance of API usability, interface specifications, and documentation quality in supporting correct software usage~\cite{dekel2009improving, de2009automatic, maalej2013patterns, piccioni2013empirical, nahar2022collaboration, subramanian2014live}. These works focus on API documentation generation~\cite{nybom2018systematic}, API documentation optimization~\cite{shi2011empirical} and documentation smell detection~\cite{khan2021automatic} to improve documentation readability, completeness, and consistency for human developers. With the rapid development of LLM agents, tool documentation has also attracted research attention. Patel et al.~\cite{play2prompt} automatically generate executable tool-use~demonstrations to reduce the manual effort of writing tool documentation, while \textsc{EasyTool}~\cite{easytool} compresses existing tool documentation into concise instructions using predefined templates to reduce context overhead. A few works~\cite{jtpro, qu2025exploration, vgco} further leverage LLMs and feedback from tool execution to iteratively refine the tool documentation for better tool invocation correctness. 

These studies improve tool documentation quality and the tool-use capabilities of LLM agents to some extent. However, they mainly optimize tool documentation from an isolated, tool-level perspective, focusing on rewriting, correction, compression, or refinement within existing information fields. In contrast, \tool does not treat documentation of each tool as an independent optimization target. Instead, it considers the interactions among tools within the same task domain and transfers optimization experience derived from domain-level and agent-aware execution failures to improve individual tool documentation by adding, removing, or refining information fields. In this way, \tool aims not only to enhance the invocation correctness of each tool, but also to improve the overall task success of different LLM agents.

\section{Conclusion}\label{sec:conclusion}
We conduct an empirical study of tool documentation for LLM agents, revealing the varying effectiveness of different information fields on task success rate across task domains, LLM backbones, and agent paradigms. Our findings~demonstrate that fixed tool documentation cannot consistently support~different agent settings. Thus, we propose \tool,~an adaptive tool documentation optimization framework that automatically refines tool documentation based on the execution traces of LLM agents. Large-scale experiments have been conducted to demonstrate the effectiveness and efficiency of \tool.



{\footnotesize
\bibliographystyle{IEEEtranS}
\bibliography{IEEEabrv,src/reference}
}

\end{document}